\documentclass[10pt,twocolumn,letterpaper]{article}

\usepackage[pagenumbers]{cvpr} 

\definecolor{cvprblue}{rgb}{0.21,0.49,0.74}
\usepackage[pagebackref,breaklinks,colorlinks,allcolors=cvprblue]{hyperref}
\usepackage{multirow}
\usepackage{soul}
\usepackage[accsupp]{axessibility}  

\def\paperID{} 
\def\confName{CVPR}
\def\confYear{2025}

\title{Depth Any Camera: Zero-Shot Metric Depth Estimation from Any Camera}

\author{
{Yuliang Guo$^{1}$\thanks{Corresponding author.}}\,\,\thanks{Equal technical contribution.},\,\, Sparsh Garg$^{2}$\footnotemark[2],\,\, S. Mahdi H. Miangoleh$^{3}$,\,\,  Xinyu Huang$^{1}$,\,\, Liu Ren$^{1}$
    \\
{\small$^{1}$Bosch Research North America \& Bosch Center for Artificial Intelligence (BCAI)}\\
{\small$^{2}$Carnegie Mellon University \quad $^{3}$Simon Fraser University}\\
{\small\tt$^{1}$[yuliang.guo2,xingyu.huang,liu.ren]@us.bosch.com}\\
{\small\tt$^{2}$sparshg@andrew.cmu.edu \quad $^{3}$smh31@sfu.ca}\\
{\small\url{https://yuliangguo.github.io/depth-any-camera}}
}

\begin{document}
\maketitle

\begin{abstract}

While recent depth foundation models exhibit strong zero-shot generalization, achieving accurate metric depth across diverse camera types—particularly those with large fields of view (FoV) such as fisheye and 360-degree cameras—remains a significant challenge. This paper presents Depth Any Camera (DAC), a powerful zero-shot metric depth estimation framework that extends a perspective-trained model to effectively handle cameras with varying FoVs. The framework is designed to ensure that all existing 3D data can be leveraged, regardless of the specific camera types used in new applications. Remarkably, DAC is trained exclusively on perspective images but generalizes seamlessly to fisheye and 360-degree cameras without the need for specialized training data. DAC employs Equi-Rectangular Projection (ERP) as a unified image representation, enabling consistent processing of images with diverse FoVs. Its core components include Pitch-aware Image-to-ERP conversion with efficient online augmentation to simulate distorted ERP patches from undistorted inputs, FoV alignment operations to enable effective training across a wide range of FoVs, and multi-resolution data augmentation to further address resolution disparities between training and testing. DAC achieves state-of-the-art zero-shot metric depth estimation, improving $\delta_1$ accuracy by up to 50\% on multiple fisheye and 360-degree datasets compared to prior metric depth foundation models, demonstrating robust generalization across camera types.

\end{abstract}


\section{Introduction}
\label{sec:intro}

\begin{figure}[!t]
    \centering
    \includegraphics[width=0.45\textwidth]{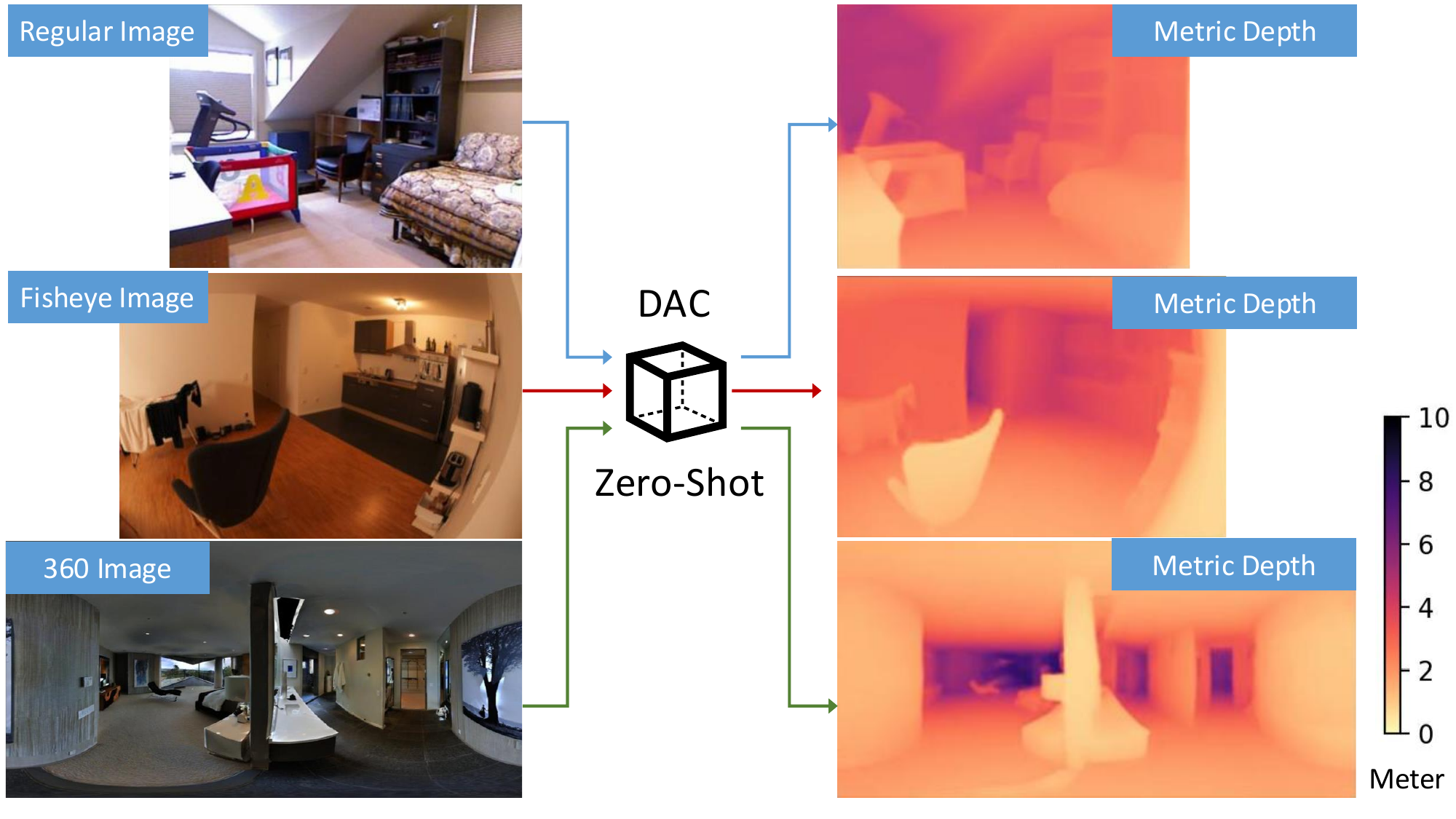}
    \caption{We introduce Depth Any Camera (DAC) framework, which leverages large-scale datasets containing perspective camera images to train a single depth estimation model capable of conducting zero-shot metric depth estimation on images captured various types of cameras, including those captured from large FoV fisheye and $360^\circ$ cameras.
    }
   \label{fig:teasor}
   \vspace{-4mm}
\end{figure}

Depth estimation from monocular cameras is a foundational challenge for applications like autonomous driving, AR/VR, and robotics. While early deep learning methods relied on supervised training using single datasets and depth sensor \cite{keselman2017intelrealsensestereoscopicdepth} supervision \cite{journals/corr/LeeHKS2019/bts,conf/iccv/RanftlBK21/dpt,conf/cvpr/BhatAW21/adabins}, monocular depth estimation remains challenging due to scale-depth ambiguity. Expanding training datasets has been a key strategy to enhance robustness, with self-supervised approaches using sequential frames~\cite{conf/iccv/GodardAFB19/monodepth2,watson2021temporalopportunistselfsupervisedmultiframe,conf/cvpr/WuWHNS22}. However, these methods often underperform due to self-supervision ambiguity, view inconsistencies and dynamic objects. Recent methods, such as MiDaS~\cite{journals/pami/RanftlLHSK22/midas}, leverage large-scale datasets with 3D supervision, normalizing scale differences across datasets to enable zero-shot testing. However, they primarily provide relative depth rather than metric depth.

Recent methods tackle zero-shot metric depth estimation by addressing the challenges of inconsistent scaling factors in depth ground truth caused by varying camera intrinsic parameters. Several works demonstrate impressive generalization capabilities on novel images~\cite{journals/corr/BhatBWWM23/zoedepth,conf/cvpr/YangKHXFZ24/depthanything,conf/iccv/000600CYWCS23/metric3d,journals/corr/abs-2404-15506/metric3dv2,conf/cvpr/PiccinelliYSSLG24/unidepth}, establishing themselves as foundational depth models for downstream tasks. However, these approaches often struggle with large field-of-view (FoV) cameras like fisheye and $360^\circ$ cameras, where performance significantly declines compared to standard perspective cameras.

\begin{figure}[!thb]
    \centering
    \includegraphics[width=\linewidth]{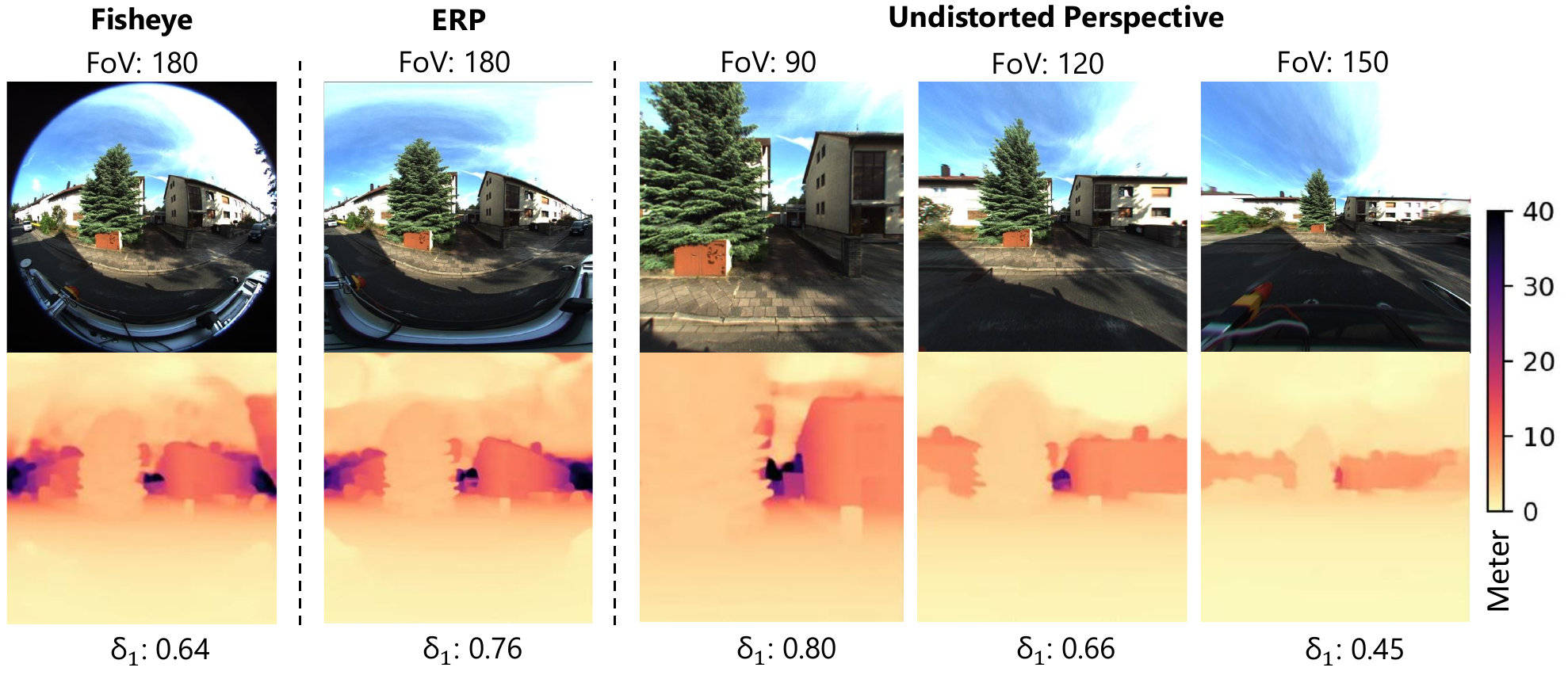}
    \caption{\textbf{Challenges on zero-shot test on large FoV camera images.} Metric depth estimation models trained on perspective images (e.g., Metric3Dv2~\cite{journals/corr/abs-2404-15506/metric3dv2}) experience significant performance degradation when applied to fisheye images. Degradation is less pronounced when using an undistorted portion with a highly limited FoV or its ERP conversion.
    }
   \label{fig:issue:fisheye}
   \vspace{-4mm}
\end{figure}

As illustrated in Fig.~\ref{fig:issue:fisheye}, large FoV images can be represented in multiple formats, but generating the best-performing undistorted image for perspective-based depth models often leads to substantial FoV loss. Despite these limitations, large FoV inputs are crucial for efficiency-critical downstream applications such as large-scale detection~\cite{DBLP:conf/iccv/YangLXS0W23, plaut20213dobjectdetectionsingle}, segmentation~\cite{DBLP:conf/cvpr/YogamaniUNK22, ye2020universalsemanticsegmentationfisheye}, SLAM~\cite{DBLP:conf/cvpr/GallagherSHM23,DBLP:conf/accv/WangCLLGLC18, zhang2023bamfslambundleadjustedmultifisheye, wang2019cubemapslampiecewisepinholemonocularfisheye}, interactive 3D scene generation~\cite{DBLP:journals/corr/abs-2406-09394}, and robotic demonstration capturing~\cite{DBLP:journals/corr/abs-2402-10329,DBLP:journals/corr/abs-2409-19499,grauman2024ego}.

Achieving zero-shot depth generalization across any FoV camera presents several challenges: (1) selecting a unified camera model to represent diverse FoVs, (2) effectively leveraging perspective training datasets to generalize to data spaces observable only from large FoV cameras, (3) managing drastically different training image sizes in the unified space caused by varying FoVs, and (4) handling resolution inconsistencies between training and testing phases.

In this paper, we present \textbf{Depth Any Camera (DAC)}, a novel zero-shot metric depth estimation framework that enables a depth model trained exclusively on perspective images to generalize across cameras with widely varying FoVs, including fisheye and $360^\circ$ cameras (see Fig.~\ref{fig:teasor}). DAC employs Equi-Rectangular Projection (ERP) as a canonical representation to unify images from diverse FoVs into a shared space. 
A key innovation is the introduction of an efficient \textbf{Pitch-aware Image-to-ERP conversion} based on grid sampling and Gnomonic Geometry~\cite{weisstein:gnomonic}, enabling seamless ERP-space data augmentations. Specifically, pitch-aware ERP conversion with pitch-angle augmentation projects perspective images into high-distortion regions of the ERP space, effectively simulating observations unique to large-FoV cameras. This enhances DAC’s zero-shot generalization, allowing it to extrapolate beyond the perspective domain to a broader range of camera types.
To facilitate learning from mixed datasets, we propose a \textbf{FoV alignment} process that normalizes diverse-FoV training samples to a predefined ERP patch size, preserving content while minimizing computational overhead.
Additionally, multi-resolution augmentation is applied to address resolution mismatches, allowing the model to learn scale-equivariant features and adapt to a flexible range of testing resolutions. In summary, our contributions are as follows:

\begin{itemize} 
    \item We propose a novel zero-shot metric depth estimation framework capable of handling images from any camera type, including fisheye and $360^\circ$ images, using a model trained exclusively on perspective data. 
    \item We introduce an efficient pitch-aware Image-to-ERP conversion that simulates the high-distortion characteristics of large-FoV cameras from perspective inputs, enhancing zero-shot generalization.
    \item We develop a FoV alignment process that enables effective training across cameras with diverse FoVs within a unified ERP space, along with a multi-resolution training strategy to address resolution mismatches between training ERP patches and testing images.
    \item Our method achieves State-of-The-Art (SoTA) zero-shot performance on all large FoV testing datasets, delivering up to a 50\% improvement in $\delta_1$ accuracy on indoor fisheye and $360^\circ$ datasets, showcasing strong generalization across diverse camera types.
\end{itemize}

\section{Related Works}
\label{sec:related:work}

\subsection{Zero-Shot Monocular Depth Estimation}
Recent approaches to zero-shot metric depth estimation tackle the challenge of inconsistent scaling factors in depth ground truth due to varying camera intrinsic parameters \cite{conf/iccv/GuiziliniV0AG23/scaleprior,journals/corr/BhatBWWM23/zoedepth,conf/cvpr/YangKHXFZ24/depthanything, yang2024depthv2,conf/iccv/000600CYWCS23/metric3d,journals/corr/abs-2404-15506/metric3dv2,conf/cvpr/PiccinelliYSSLG24/unidepth}. Zoedepth~\cite{journals/corr/BhatBWWM23/zoedepth} introduces an advanced network architecture, while DepthAnything~\cite{conf/cvpr/YangKHXFZ24/depthanything, yang2024depthv2} employs a sophisticated unsupervised learning paradigm. However, their performance in metric depth estimation is limited without tackling inconsistency camera intrinsics. Metric3D~\cite{conf/iccv/000600CYWCS23/metric3d,journals/corr/abs-2404-15506/metric3dv2} and UniDepth~\cite{conf/cvpr/PiccinelliYSSLG24/unidepth} address scaling inconsistencies by converting images into a canonical camera space. Metric3D uses intrinsic parameters for this preprocessing, whereas UniDepth incorporates a network branch to estimate and convert intrinsics on the fly. Despite these advances, none of these methods achieve satisfactory zero-shot performance on large FoV images, presenting unique challenges in unifying diverse FoVs and supporting effective model learning.

\subsection{Depth Estimation from Large FoV Cameras}
Depth estimation for fisheye, $360^\circ$ cameras has grown in popularity, as large FoVs capture richer contextual information that enhances depth estimation~\cite{journals/ral/JiangSZDH21/unifuse,conf/cvpr/LiGY0DR22/omnifusion,conf/eccv/ShenLLNZZ22/panoformer,conf/iccv/YunSLLR23/egformer,conf/cvpr/AiCCSW23/hrdfuse}. A key challenge for these cameras is managing position-dependent distortions, which vary by camera models. Approaches to address this include deformable CNNs~\cite{conf/nips/SuG17/deformcnn360, zhu2019deformableconvnetsv2deformable,xiong2024efficient}, which adapt kernel shapes to compensate for distortions, as well as methods that segment ERP images to reduce distortion effects before merging~\cite{journals/ral/JiangSZDH21/unifuse,rey2022360monodepth}.
More recent methods leverage transformers to handle these distortions~\cite{conf/cvpr/LiGY0DR22/omnifusion,conf/eccv/ShenLLNZZ22/panoformer,conf/iccv/YunSLLR23/egformer, feng2023simfirsimpleframeworkfisheye}.
While transformer-based networks have improved in-domain performance, they are approaching saturation, indicating that distortion is not the only challenge. Instead, the lack of large-scale FoV-specific training data is a key bottleneck for generalization. No current methods enable an unified depth estimation model trained on mixed large-scale perspective datasets to achieve zero-shot generalization on ERP or fisheye images.


\section{Notations and Preliminaries}

\begin{figure}[!tb]
    \centering
    \includegraphics[width=0.47\textwidth]{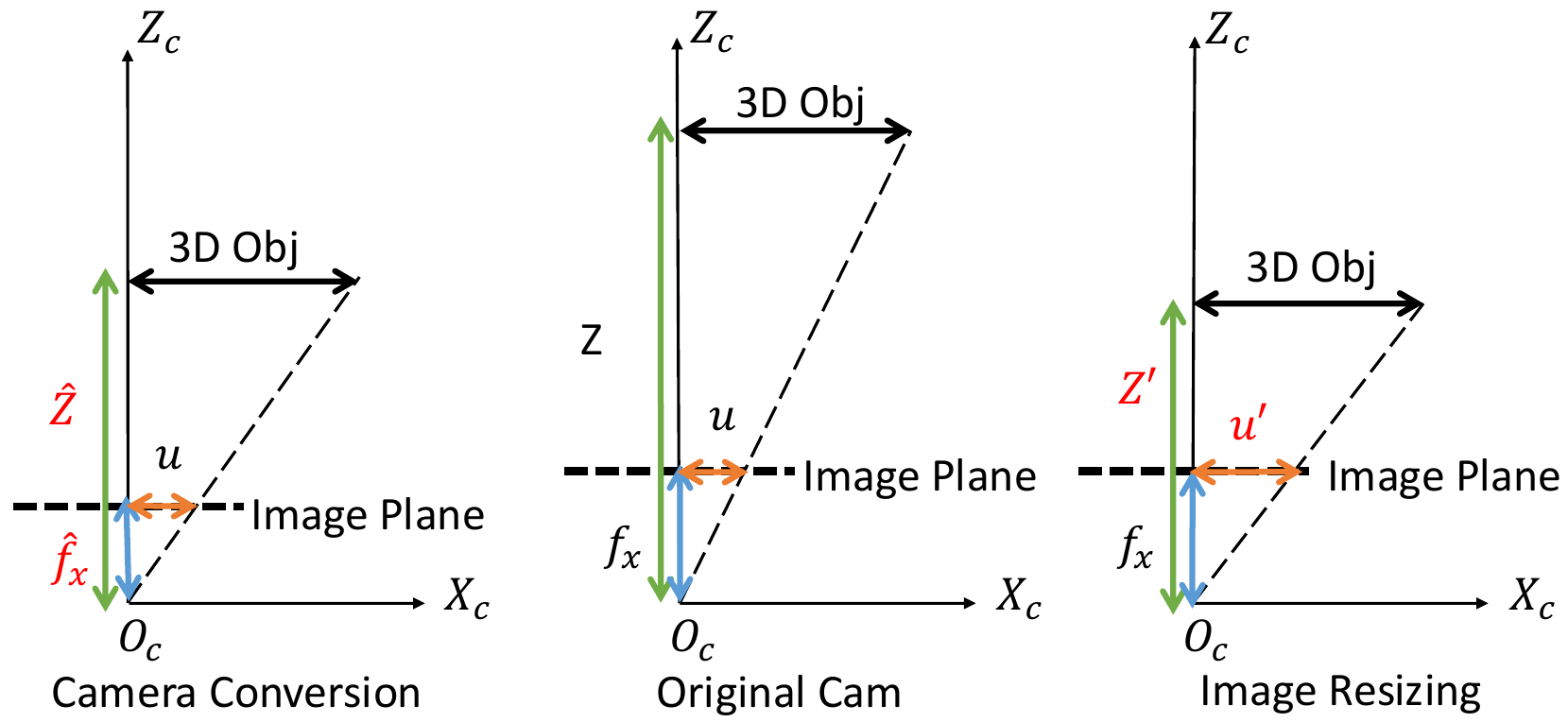}
    \caption{\textbf{Depth Scaling in Canonical Model Conversion and Image Resizing.} The apparent 2D size of an object \( u \) in an image depends on its 3D dimensions \( X \), depth \( Z \), and camera focal length \( f_x \). \textit{Left}: Converting a perspective camera model to a canonical model with a different focal length \( \hat{f}_x \) requires scaling the depth values proportionally, so \( \hat{Z} = \frac{\hat{f}_x Z}{f_x} \). \textit{Center}: The original camera setup, showing the direct relationship between object size, depth, and focal length. \textit{Right}: When the camera model is fixed but the image is resized to \( u' \), this simulates viewing the same 3D object at a different distance, necessitating depth scaling for accurate metric depth, with \( Z' = \frac{u Z}{u'} \).
    }
   \label{fig:depth-scale-ops}
   \vspace{-4mm}
\end{figure}

\noindent \textbf{Depth Scaling Operations.} Monocular depth estimation is inherently ill-posed, as different 3D object sizes and depths can produce the same 2D appearance. Deep learning models rely on learning an object's 3D dimensions from its 2D appearance~\cite{conf/iccv/DijkC19,conf/iccv/GuiziliniV0AG23/scaleprior,conf/eccv/guo2024/supnerf} to infer depth, leading to the scaling operation illustrated in the right panel of Fig.~\ref{fig:depth-scale-ops}. When using mixed camera data, apparent object size also depends on focal length, making accurate 2D-to-depth mapping dependent on appropriately scaling ground-truth depths when converting a perspective model to a canonical model, as shown in the left panel of Fig.~\ref{fig:depth-scale-ops}. These scaling operations are central to the Metric3D~\cite{conf/iccv/000600CYWCS23/metric3d, journals/corr/abs-2404-15506/metric3dv2} pipeline and are integrated into our ERP-based approach.

\noindent \textbf{EquiRectangular Projection (ERP).} 
Equi-Rectangular Projection (ERP) is an image representation based on a spherical camera model, where each pixel corresponds to a specific latitude \(\lambda\) and longitude \(\phi\). A full ERP space spans \(180^\circ\) in latitude and \(360^\circ\) in longitude, making it ideal for handling diverse FoV cameras. The ERP image height is the only parameter needed to define the ERP space, allowing both training and testing images to be consistently converted into this space, regardless of the original FoV.

Transformations between standard images and ERP images use Gnomonic Projection transformation~\cite{weisstein:gnomonic}, which offers a closed-form mapping between tangent image coordinates \((x_t, y_t)\) and spherical coordinates \((\lambda, \phi)\), assuming a tangent plane centered at \((\lambda_c, \phi_c)\) of a unit sphere. Specifically, as shown in Fig.~\ref{fig:im2erp:fov:align}, this mapping is given by:
{\small
\begin{align}
    x_t &= \frac{\bar{x}}{\cos c} = \frac{\cos \phi \sin (\lambda - \lambda_c)}{\cos c} \label{eq:gp:x} \\
    y_t &= \frac{\bar{y}}{\cos c} = \frac{\cos \phi_c \sin \phi - \sin \phi_c \cos \phi \cos (\lambda - \lambda_c)}{\cos c} \label{eq:gp:y}
\end{align}
}
where $(\bar{x},\bar{y},\cos c)$ represents a point on the unit sphere, and $c$ is the angular distance between $(\lambda, \phi)$ and $(\lambda_c, \phi_c)$, can be calculated as:
{\small
\begin{equation}
    \cos c = \sin \phi_c \sin \phi + \cos \phi_c \cos \phi \cos (\lambda - \lambda_c)
\end{equation}
}
We use these transformations to enable an efficient ERP conversion and data augmentation process, creating a streamlined pipeline that supports zero-shot generalization for depth estimation across various FoV cameras.



\section{Depth Any Camera}
\label{sec:method}

\begin{figure*}[!tb]
    \centering
    \includegraphics[width=0.96\textwidth]{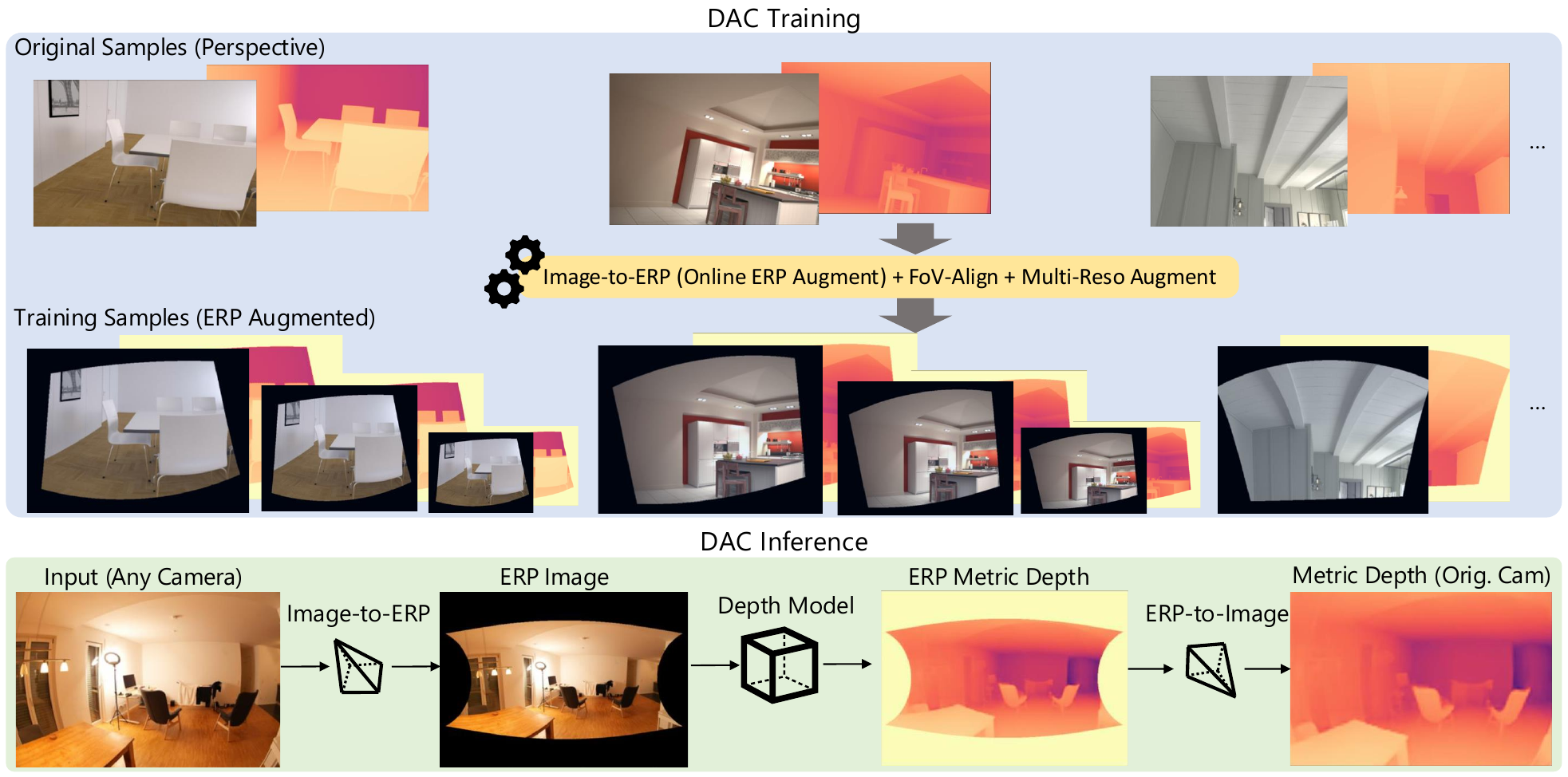}
    \caption{\textbf{Depth Any Camera Pipeline.} Our DAC framework converts data from any camera type into a canonical ERP space, enabling a model trained solely on perspective images to process large-FoV test data consistently for metric depth estimation. During training, we introduce an effective pitch-aware Image-to-ERP conversion with online data augmentation to simulate high-distortion regions unique to large-FoV images. The proposed FoV-Align process normalizes diverse-FoV data to a predefined ERP patch size, maximizing training efficiency. During inference, images from any camera type are converted into ERP space for depth estimation, with an optional step to map the ERP output back to the original image space for visualization.
    }
   \label{fig:pipeline}
   \vspace{-4mm}
\end{figure*}


We propose \textbf{Depth Any Camera (DAC)}, a depth model training framework designed to achieve zero-shot generalization across diverse camera models, including perspective, fisheye, and $360^\circ$ cameras. As illustrated in Fig. \ref{fig:pipeline}, images from different camera types and FoVs are transformed into a canonical ERP space during both the training and testing phases. For training, we leverage the extensive perspective image datasets by converting them into smaller ERP patches for efficient learning. During testing, large FoV images are similarly converted into the canonical ERP space, allowing the trained model to predict metric depth consistently, without getting confused by different camera intrinsic and distortion parameters. 

Several key components are designed to address specific challenges in implementing the DAC framework. In Sec. \ref{sec:im2erp}, we present an efficient pitch-aware Image-to-ERP conversion method that simulate large-FoV images at patch level and supports online augmentation within the ERP space. Sec. \ref{sec:fov:align} introduces a FoV alignment process, an effective data augmentation technique that maximizes content inclusion while minimizing computational waste on background areas, using a single predefined ERP patch size. In Sec. \ref{sec:multi:reso:train}, we describe a multi-resolution data augmentation approach aimed at training a transformer-based network capable of handling a broad range of testing resolutions. 

The proposed DAC framework is compatible with various depth estimation network architectures, which are not the primary focus of this paper. Without loss of generality, we employ iDisc~\cite{conf/cvpr/PiccinelliSY23/idisc} for its simplicity and effectiveness, and for its incorporation of two prototypical attention modules, namely cross-attention and self-attention. We use the SIlog loss function \cite{eigen2014depthmappredictionsingle} for training our models.

\subsection{Pitch-Aware Image-to-ERP Conversion}
\label{sec:im2erp}


An input image can be efficiently converted to its corresponding ERP patch through grid sampling combined with gnomonic projection. Assuming an ERP space with height $H_{E}$, width $W_{E} = 2 H_{E}$, and the image center at latitude $\lambda_c$, longitude $\phi_c$,  with a target ERP patch size of $H_{e} \times W_{e}$, the ERP patch coordinates $(u_{e}, v_{e})$ can be mapped to spherical coordinates as $\phi = \frac{2 \pi W_e}{W_{E}} (u_{e} - \frac{W_{e}}{2}) + \phi_c$, and $\lambda =  \frac{\pi H_e}{H_{E}} (v_{e} - \frac{H_{e}}{2}) + \lambda_c$.
Using Gnomonic Geometry presented in Eq.~\ref{eq:gp:x} and Eq.~\ref{eq:gp:y}, we obtain the corresponding normalized image coordinate $(x_t, y_t, 1)$ in the tangent plane and $(\bar{x}, \bar{y}, \cos c)$ on the unit sphere. To map this coordinate to the actual image coordinate $(u, v)$, we apply distortion and projection functions based on the given camera parameters:
{\small
\vspace{-4mm}
\begin{align}
    (x_d, y_d) &= f_d (\bar{x}, \bar{y}, \cos c, D_c) \label{eq:distortion}\\ 
    (u, v) &=  f_p(x_d, y_d, K_c)
\end{align}
}
where $f_d$ is the distortion function with distortion parameters $D_c$, and $f_p$ is the projection function with intrinsic parameters $K_c$. If the input image has no distortion, projection function is directly applied to $(x_t, y_t)$. Details on applying distortion models are included in Supplemental Sec.~\ref{sec:distortion:model}.

\begin{figure}[!tb]
    \centering
    \includegraphics[width=0.45\textwidth]{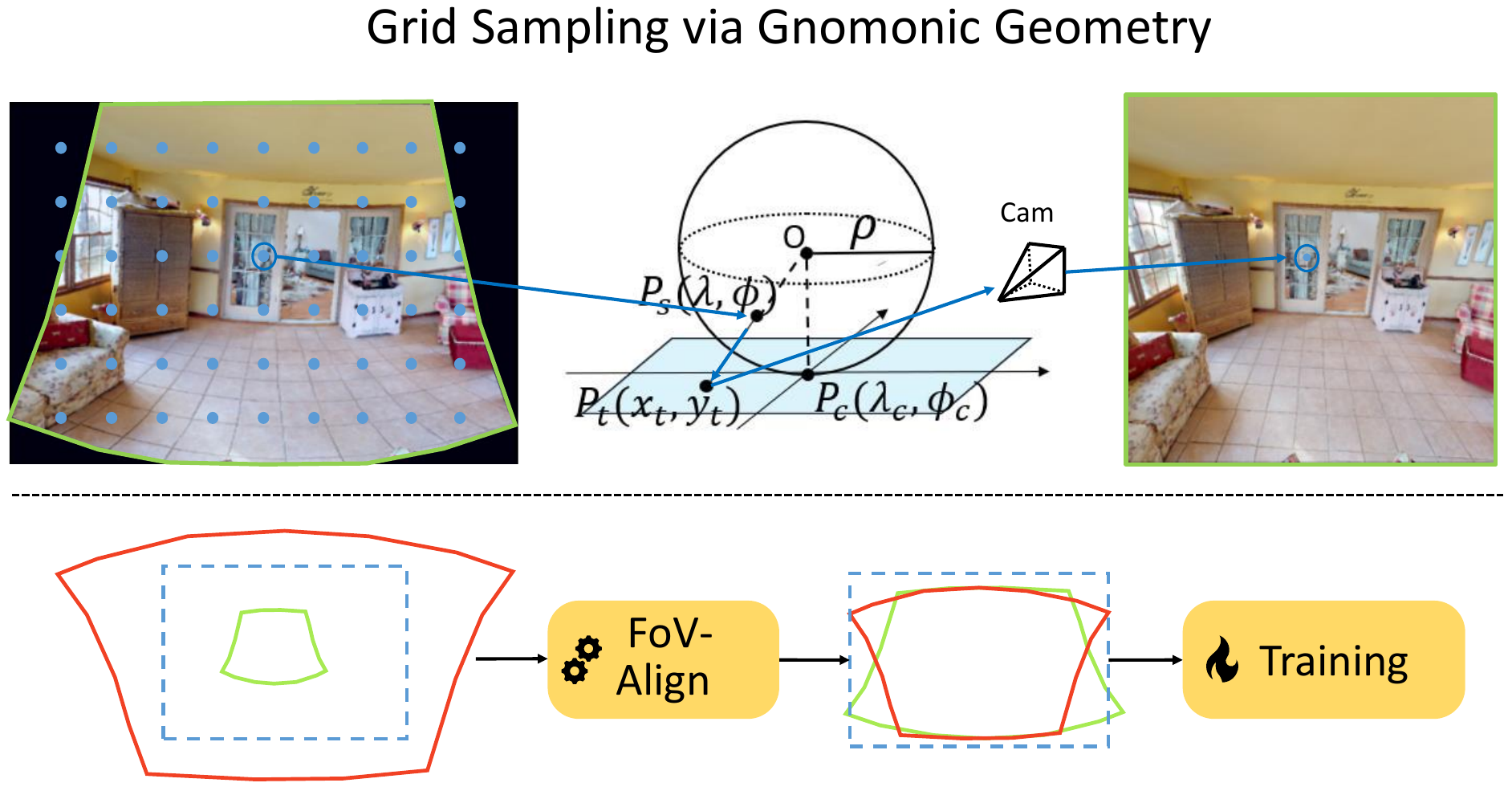}
    \caption{\textbf{Pitch-Aware ERP Conversion and FoV Alignment.}  \textit{Top}: Grid Sampling is applied for an efficient Image-to-ERP conversion. Each ERP grid sample's corresponding location in the input image is computed using gnomonic geometry and specific camera projection parameters. Given the patch center latitude $\lambda$ determined by the camera's pitch angle, it makes the converted patch to represent high-distortion regions in the ERP space. \textit{Bottom}: The FoV-Align process normalizes diverse-FoV ERP patches (shown in red and green) to match the height of a single predefined ERP patch (outlined in blue), ensuring efficient training.
    }
   \label{fig:im2erp:fov:align}
   \vspace{-4mm}
\end{figure}

As shown in Fig.~\ref{fig:im2erp:fov:align}, with uniformly sampled grid points within the target ERP patch, each grid point can be mapped directly to a corresponding location in the input image. This mapping facilitates efficient transformation of the captured image into an ERP patch via grid sampling. Essentially, each grid point in the ERP patch maps to a specific floating-point position in the input image, and its value is obtained by interpolating from the neighboring pixel values.

This ERP conversion enables a powerful training pipeline when the latitude of the tangent plane center \(\lambda_c\) is defined by the camera's pitch angle in training.
When camera orientation is available or can be estimated~\cite{DBLP:conf/cvpr/JinZHWBSF23/perspectivefield}, perspective data can be projected to various latitudes in the ERP space, enabling the simulation of regions uniquely visible from large-FoV cameras, as shown in Fig.~\ref{fig:im2erp:fov:align}, and Supplemental Fig.~\ref{fig:pitch:aware}.
This \textbf{pitch-aware conversion} is crucial for improving the generalization of trained models to previously unobserved large-FoV data, as demonstrated in Sec.~\ref{sec:ablation}, because neural networks alone have limited capacity to generalize to extrapolated data spaces~\cite{conf/iclr/XuZLDKJ21/nnextrapolate}.

Another notable advantage is the seamless ability to perform \textbf{online augmentation} efficiently in the ERP space. For datasets with limited pitch variation~\cite{journals/ijrr/GeigerLSU13/kitti,conf/cvpr/GuiziliniAPRG20/ddad,DBLP:conf/corl/HoustonZBYCJOIO20/lyft,conf/ijcai/ZamirSSGMS19/taskonomy}, a unique pitch augmentation can be efficiently applied by adding noise to \(\lambda_c\), generating ERP patches with varying shapes, as shown in Fig.~\ref{fig:pipeline}.
In addition, common augmentations, such as scaling, rotation, and translation—commonly applied to perspective images—can be directly applied to the normalized image coordinates $(x_t, y_t)$ as follows:
\begin{align}
    \begin{bmatrix}
        x_t'\\
        y_t'
    \end{bmatrix}
    =
    s_\sigma
    \begin{bmatrix}
     R_\sigma \;\; T_\sigma   
    \end{bmatrix}
    \begin{bmatrix}
        x_t\\
        y_t
    \end{bmatrix}
    \label{eq:aug}
\end{align}
where $s_\sigma$ is a scale factor, $R_\sigma$ is 2D rotation matrix, and $T_\sigma$ is a 2D translation vector corresponding to the applied augmentations. 


\subsection{FoV Alignment}
\label{sec:fov:align}

When training data include a wide range of camera FoVs within perspective images, such as in the HM3D~\cite{ramakrishnan2021/hm3d} dataset produced by OmniData~\cite{conf/iccv/EftekharSMZ21/omnidata}, the corresponding ERP regions can vary significantly in size, as shown in Fig.~\ref{fig:im2erp:fov:align}. There is no single crop size that can consistently capture most content information for certain images without wasting substantial computation on background padding for others. This creates a dilemma in prioritizing between training efficiency and richness of information, and it can also reduce training quality when samples exhibit drastically different content-to-background ratios.

To address this challenge, we introduce a simple yet effective FoV Alignment operation that adjusts the FoV of each input image to match the predetermined crop area FoV. Specifically, this process applies a specific scaling augmentation, as described in Sec.~\ref{sec:im2erp} and Eq.~\ref{eq:aug}, specifically $ \text{Fov}_e = \frac{H_e \pi}{H_E} \;\; \text{and} \;\; s_\sigma = \frac{\text{Fov}_c}{\text{Fov}_e}$,
where $\text{Fov}_c$ is derived from actual camera parameters, and $\text{Fov}_e$ is ERP patch’s vertical FoV. As illustrated in Fig.~\ref{fig:im2erp:fov:align}, this approach allows a single predefined ERP patch size to maximize the inclusion of relevant content and minimize computational waste on background, making it ideal for an efficient training pipeline.


\subsection{Multi-Resolution Training}
\label{sec:multi:reso:train}


Training ERP patches and testing images may prefer inconsistent resolutions for various reasons, e.g. drastically different aspect ratio, edge device limitation. 
When testing resolutions differ from the training patch size, model performance can degrade significantly, particularly with attention modules that aggregate different numbers of image tokens.

To address this issue, we adopt a multi-resolution training scheme. As illustrated in Fig.~\ref{fig:pipeline}, each ERP patch is resized to two additional lower resolutions (typically 0.7 and 0.4 of the original) to incorporate varied image resolutions in training. 
The training feeds the model three batches of images at different resolutions and sums the losses.

\begin{table*}[t!]
\caption{\textbf{Overview of Datasets.} This table summarizes the training and testing datasets used in this work. The training datasets span a range of FoVs, pitch angles, and image quality, each potentially impacting performance on different test datasets in varying degrees.} 
\vspace{-5mm}
\begin{center}
\begin{footnotesize}
\setlength\tabcolsep{0.1cm}
\begin{tabular}{l  lllllll | l lll}
\hline
\textbf{Train Dataset}                                         & \textbf{\# Imgs}       & \textbf{Scene}       & \textbf{xFoV (deg.)}                   & \textbf{\# Cams}  & \textbf{Pitch (deg.)}              & \textbf{Source}       & \textbf{Img Qual.}   & \textbf{Test Dataset}                                   & \textbf{Cam Type} & \textbf{xFoV (deg.)}  & \textbf{Scene}\\
\hline
HM3D-tiny~\cite{ramakrishnan2021/hm3d}                      &   310K        &  Indoor     &  $36^\circ - 124^\circ$       & 10K+     & $3\sigma=75^\circ$        &  Recon          &   Low    & Matterport3D \cite{Matterport3D}                    & ERP  & $360^\circ$ & Indoor   \\
Taskonomy-tiny~\cite{conf/ijcai/ZamirSSGMS19/taskonomy}     &   300K        &  Indoor     &  $45^\circ - 75^\circ$        & 10K+     & $3\sigma=24^\circ$        &  Real         &   High    & Pano3D-GV2 \cite{conf/cvpr/AlbanisZDGSAZD2/pano3d}  & ERP  & $360^\circ$ & Indoor   \\
Hypersim ~\cite{conf/iccv/RobertsRRK0PWS21/hypersim}        &   54K         &  Indoor     &  $60^\circ$   & 1        & $3\sigma=60^\circ$        &  Sim        &   High+     & ScanNet++ \cite{conf/iccv/YeshwanthLND23/scannet++} & Fisheye  & $150^\circ$ &  Indoor\\
DDAD~\cite{conf/cvpr/GuiziliniAPRG20/ddad}                  &   80K         &  Outdoor    &  $45^\circ - 60^\circ$                &  36+     &  $3\sigma \sim 10^\circ$  &  Real         &   High    & KITTI360 \cite{journals/pami/LiaoXG23/kitti360}     & Fisheye  & $180^\circ$ &  Outdoor \\
LYFT~\cite{DBLP:conf/corl/HoustonZBYCJOIO20/lyft}           &   50K         &  Outdoor    &  $20^\circ - 60^\circ$                & 6+       &  $3\sigma \sim 10^\circ$  &  Real         &   High    & & &   \\
\hline
\end{tabular}
\end{footnotesize}
\end{center}
\label{tab:dataset:overview}
\end{table*}

\begin{table*}[t!]
\caption{\textbf{Zero-Shot Test on $360^\circ$ and Fisheye Datasets.} DAC is compared with SoTA metric depth models across four large-FoV datasets. 
}
\vspace{-5mm}
\begin{center}
\begin{footnotesize}
\setlength\tabcolsep{0.1cm}
\begin{tabular}{l | l | l | l | l l l l  l  l } 
\hline 
\textbf{Test Dataset} & \textbf{Methods} & \textbf{Train Dataset} & \textbf{Backbone} & $\pmb{\delta_1}$  $\uparrow$ & $\pmb{\delta_2}$  $\uparrow$ & $\pmb{\delta_3}$  $\uparrow$  & \textbf{Abs Rel}$\downarrow$ & \textbf{RMSE}$\downarrow$ & \textbf{log10}$\downarrow$  \\
\hline
\multirow{5}{*}{Matterport3D \cite{Matterport3D}} 
 & UniDepth~\cite{conf/cvpr/PiccinelliYSSLG24/unidepth}         & Mix 3M      & ViT-L~\cite{conf/iclr/DosovitskiyB0WZ21/vit}     & 0.2576  & 0.5114 & 0.7091 & 0.7648 & 1.3827 & 0.2208\\
 & Metric3Dv2~\cite{journals/corr/abs-2404-15506/metric3dv2}    & Mix 16M     & Dinov2~\cite{journals/tmlr/OquabDMVSKFHMEA24/dinov2}    & 0.4381  & 0.7311 & 0.8735 & 0.2924 & 0.8842 & 0.1546\\
 & Metric3Dv2~\cite{journals/corr/abs-2404-15506/metric3dv2}    & Indoor 670K & Dinov2~\cite{journals/tmlr/OquabDMVSKFHMEA24/dinov2}    & 0.4287  & 0.7854 & 0.9333 & 0.2788 & 0.8961 & 0.1352\\
 & iDisc~\cite{conf/cvpr/PiccinelliSY23/idisc}                  & Indoor 670K & Resnet101~\cite{conf/cvpr/HeZRS16/resnet} & 0.5287 & 0.8260 & 0.9398 & 0.2757 & 0.7771 & 0.1147\\ 
 & \textbf{DAC (Ours)}                       & Indoor 670K & Resnet101~\cite{conf/cvpr/HeZRS16/resnet} & \textbf{0.7727} & \textbf{0.9562} & \textbf{0.9822} & \textbf{0.156} & \textbf{0.6185} & \textbf{0.0707}\\
\hline
\multirow{5}{*}{Pano3D-GV2 \cite{conf/cvpr/AlbanisZDGSAZD2/pano3d}}  
 & UniDepth~\cite{conf/cvpr/PiccinelliYSSLG24/unidepth}         & Mix 3M      & ViT-L~\cite{conf/iclr/DosovitskiyB0WZ21/vit}     & 0.2469  & 0.4977 & 0.7084 & 0.7892 & 1.2681 & 0.2231\\
 & Metric3Dv2~\cite{journals/corr/abs-2404-15506/metric3dv2}    & Mix 16M         & Dinov2~\cite{journals/tmlr/OquabDMVSKFHMEA24/dinov2}    & 0.4040  & 0.6929 & 0.8499 & 0.3070 & 0.8549 & 0.1664\\
 & Metric3Dv2~\cite{journals/corr/abs-2404-15506/metric3dv2}    & Indoor 670K & Dinov2~\cite{journals/tmlr/OquabDMVSKFHMEA24/dinov2}    & 0.5060  & 0.8176 & 0.9360 & 0.2608 & 0.7248 & 0.1201\\
 & iDisc~\cite{conf/cvpr/PiccinelliSY23/idisc}                  & Indoor 670K & Resnet101~\cite{conf/cvpr/HeZRS16/resnet} & 0.5629 & 0.8222 & 0.9332 & 0.2657 & 0.6446 & 0.1122\\ 
 & \textbf{DAC (Ours)}                      & Indoor 670K & Resnet101~\cite{conf/cvpr/HeZRS16/resnet} & \textbf{0.8115} & \textbf{0.9549} & \textbf{0.9860} & \textbf{0.1387} & \textbf{0.4780} & \textbf{0.0623}\\
\hline
\multirow{5}{*}{ScanNet++ \cite{conf/iccv/YeshwanthLND23/scannet++}} 
 & UniDepth~\cite{conf/cvpr/PiccinelliYSSLG24/unidepth}         & Mix 3M      & ViT-L~\cite{conf/iclr/DosovitskiyB0WZ21/vit}     & 0.3638  & 0.6461 & 0.8358 & 0.4971 & 1.1659 & 0.1648\\
 & Metric3Dv2~\cite{journals/corr/abs-2404-15506/metric3dv2}    & Mix 16M     & Dinov2~\cite{journals/tmlr/OquabDMVSKFHMEA24/dinov2}    & 0.5360  & 0.8218 & 0.9350 & 0.2229 & 0.8950 & 0.1177\\
 & Metric3Dv2~\cite{journals/corr/abs-2404-15506/metric3dv2}    & Indoor 670K & Dinov2~\cite{journals/tmlr/OquabDMVSKFHMEA24/dinov2}    & 0.6489  & 0.8920 & 0.9558 & 0.1915 & 0.9779 & 0.0938\\
 & iDisc~\cite{conf/cvpr/PiccinelliSY23/idisc}                  & Indoor 670K & Resnet101~\cite{conf/cvpr/HeZRS16/resnet} & 0.6150  & 0.8780 & 0.9617 & 0.2712 & 0.4835 & 0.0972\\
 & \textbf{DAC (Ours)}                      & Indoor 670K & Resnet101~\cite{conf/cvpr/HeZRS16/resnet} & \textbf{0.8517}  & \textbf{0.9693} & \textbf{0.9922} & \textbf{0.1323} & \textbf{0.3086} & \textbf{0.0532}\\
\hline
\multirow{5}{*}{KITTI360 \cite{journals/pami/LiaoXG23/kitti360}} 
 & UniDepth~\cite{conf/cvpr/PiccinelliYSSLG24/unidepth}         & Mix 3M       & ViT-L~\cite{conf/iclr/DosovitskiyB0WZ21/vit}     & 0.4810  & 0.8397 & 0.9406 & 0.2939 & 6.5642 & 0.1221\\
 & Metric3Dv2~\cite{journals/corr/abs-2404-15506/metric3dv2}    & Mix 16M      & Dinov2~\cite{journals/tmlr/OquabDMVSKFHMEA24/dinov2}    & 0.7159  & 0.9323 & 0.9771 & 0.1997 & 4.5769 & 0.0811\\
 & Metric3Dv2~\cite{journals/corr/abs-2404-15506/metric3dv2}    & Outdoor 130K & Dinov2~\cite{journals/tmlr/OquabDMVSKFHMEA24/dinov2}    & 0.7675  & 0.9370 & 0.9756 & 0.1521 & 4.6610 & 0.0723\\
 & iDisc~\cite{conf/cvpr/PiccinelliSY23/idisc}                  & Outdoor 130K & Resnet101~\cite{conf/cvpr/HeZRS16/resnet} & 0.7833  & 0.9384 & 0.9753 & 0.1598 & 4.9122 & 0.0704\\
 & \textbf{DAC (Ours)}                      & Outdoor 130K & Resnet101~\cite{conf/cvpr/HeZRS16/resnet} & \textbf{0.7858}  & \textbf{0.9388} & \textbf{0.9775} & \textbf{0.1559} & \textbf{4.3614} & \textbf{0.0684}\\
\hline

\end{tabular}
\end{footnotesize}
\end{center}
\label{tab:exp:main}
\vspace{-4mm}
\end{table*}

\section{Experiments}
\label{sec:exp}

\subsection{Experimental Setup}

\noindent\textbf{In-Domain Training Datasets.} For indoor experiments, we use Habitat-Matterport 3D (HM3D)~\cite{ramakrishnan2021/hm3d}, Taskonomy~\cite{conf/ijcai/ZamirSSGMS19/taskonomy}, and Hypersim~\cite{conf/iccv/RobertsRRK0PWS21/hypersim}, totaling 670K perspective images with distinct characteristics. To streamline training, we use the first 50 scenes from HM3D and Taskonomy (tiny versions provided by OmniData~\cite{conf/iccv/EftekharSMZ21/omnidata}). For outdoor data, we use DDAD~\cite{conf/cvpr/GuiziliniAPRG20/ddad} and LYFT~\cite{DBLP:conf/corl/HoustonZBYCJOIO20/lyft}, totaling 130K images. Table~\ref{tab:dataset:overview} summarizes the datasets in use, showing varying distributions in FoV, pitch angles, sources, and image quality.

\noindent\textbf{Zero-Shot Testing Datasets.} We evaluate DAC on two $360^\circ$ datasets—Matterport3D~\cite{Matterport3D} and Pano3D-GV2~\cite{conf/cvpr/AlbanisZDGSAZD2/pano3d}—and two fisheye datasets—ScanNet++~\cite{conf/iccv/YeshwanthLND23/scannet++} and KITTI360~\cite{journals/pami/LiaoXG23/kitti360}—all featuring larger FoVs than perspective images. Our primary experiments focus on these datasets to assess zero-shot generalization to large FoV cameras. Additional evaluations on NYUv2~\cite{Silberman:ECCV12:nyuv2} and KITTI~\cite{journals/ijrr/GeigerLSU13/kitti} are provided in the supplementary material, demonstrating DAC's performance on perspective data relative to SoTA methods.

\noindent\textbf{Evaluation Details.} We assess DAC’s generalization to large FoV cameras across both indoor and outdoor scenes, training separate models for each without data mixing to simplify training. For fair comparison, competing models are either re-trained with the same data splits or use their largest versions trained on extensive datasets. Evaluations are conducted using metric depth metrics: ${\delta_1}$ $\uparrow$, ${\delta_2}$ $\uparrow$, ${\delta_3}$ $\uparrow$, Abs Rel$\downarrow$, RMSE$\downarrow$, and log10$\downarrow$.

\noindent\textbf{Baselines.} We compare DAC with the following baselines:
\begin{itemize}
    \item Metric3Dv2~\cite{journals/corr/abs-2404-15506/metric3dv2}: A SoTA foundation model in zero-shot metric depth estimation, built upon perspective canonical camera model to standardize datasets.
    \item UniDepth~\cite{conf/cvpr/PiccinelliYSSLG24/unidepth}: A more recent SoTA foundation depth model leveraging network designs to handle diverse camera parameters. We test its ability to handle large FoV cameras not included in training.
    \item iDisc~\cite{conf/cvpr/PiccinelliSY23/idisc}: Selected as a network baseline due to its use of self-attention and cross-attention modules in a straightforward yet effective network, and strong in-domain performance. As iDisc alone does not handle mixed camera parameters, we train it with Metric3Dv2~\cite{journals/corr/abs-2404-15506/metric3dv2}, and compare it to the same network trained with ours.
\end{itemize}

\noindent\textbf{Implementation Details.} In the DAC training pipeline, we set the full ERP height to \(H_{\text{erp}}=1400\) pixels, with an ERP patch size of \(500 \times 700\) pixels for both indoor and outdoor models. We use $10^\circ$ latitude augmentations for both and additionally $10^\circ$ rotation augmentation for indoor. When training the iDisc~\cite{conf/cvpr/PiccinelliSY23/idisc} model using the Metric3D~\cite{conf/iccv/000600CYWCS23/metric3d} pipeline, we use canonical focal lengths of \(f_{\text{cano}} = 519\) (NYU dataset~\cite{Silberman:ECCV12:nyuv2}) for indoor models and \(f_{\text{cano}} = 721\) (KITTI dataset~\cite{journals/ijrr/GeigerLSU13/kitti}) for outdoor models.

To test perspective models on ERP and fisheye images, specific adjustments are required. For \(360^\circ\) (ERP) images, which lack a defined focal length, we calculate a virtual focal length \(f_{\text{virtual}}\) based on pixels per latitude degree: \( \frac{1}{f_{\text{virtual}}} = \tan\left(\frac{\pi}{H_{\text{erp}}}\right) \), scaling the predicted depth with \( \frac{f_{\text{cano}}}{f_{\text{virtual}}} \) for ground-truth alignment. For fisheye images, aligning \(f_{\text{cano}}\) with the post-distortion focal length introduces significant errors, so we first convert fisheye images to ERP space and apply \( \frac{f_{\text{cano}}}{f_{\text{virtual}}} \) for metric depth evaluation.

For testing resolution, if the original resolution is less than twice the training resolution, we use it directly; for larger resolutions, we maintain the aspect ratio and align it with the training resolution. Based on this rule, we evaluate Matterport3D~\cite{Matterport3D} and Pano3D-GV2~\cite{conf/cvpr/AlbanisZDGSAZD2/pano3d} at \(512 \times 1024\), ScanNet++~\cite{conf/iccv/YeshwanthLND23/scannet++} at \(500 \times 750\), and KITTI360~\cite{journals/pami/LiaoXG23/kitti360} at \(700 \times 700\). For competing methods that are not adaptable to inconsistent resolutions compared to training, we report the higher score obtained from the two settings to ensure fairness.

In experiments, ResNet101~\cite{conf/cvpr/HeZRS16/resnet}-backbone models are trained for 60k iterations with a batch size of 48, while Swin-L~\cite{conf/iccv/LiuL00W0LG21/swintrans} and DINOv2~\cite{journals/tmlr/OquabDMVSKFHMEA24/dinov2} models are trained for 120k iterations with a batch size of 48. Finally, to support all FoV types, depth is represented as \textit{Euclidean Distance} from the camera center rather than \textit{Z-buffer} format, as the latter is incompatible with spherical projections and would yield inaccurate low depth values for fisheye or ERP images.

\subsection{Comparison with the SoTA}

\begin{figure*}[!tb]
    \centering
    \includegraphics[width=0.95\textwidth]{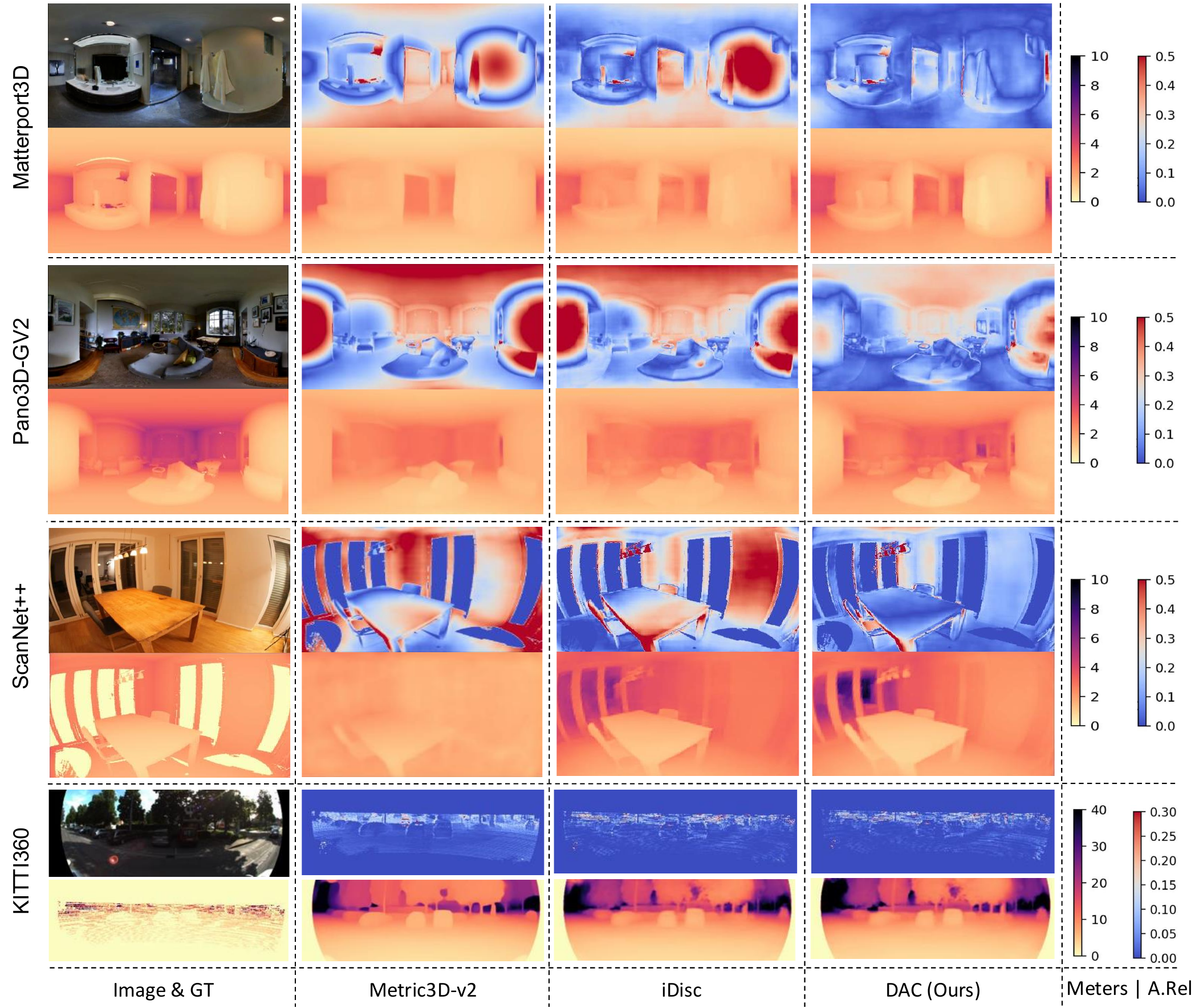}
    \caption{\textbf{Zero-Shot Qualitative Results.} For each dataset, an example is presented in two consecutive rows. The left column shows the original image and Ground-Truth depth map, followed by results from various methods. For each method, the top row displays the A.Rel map \( \downarrow \) and the bottom row shows the predicted depth map. The color range for depth and A.Rel maps is indicated in the last column.
    }
   \label{fig:vis:main}
\end{figure*}

In this section, we compare DAC with primary baselines in zero-shot generalization tests on large FoV datasets, with quantitative results reported in Table~\ref{tab:exp:main}, and qualitative results shown in Fig.~\ref{fig:vis:main}. In indoor experiments, DAC significantly outperforms pre-trained models UniDepth~\cite{conf/cvpr/PiccinelliYSSLG24/unidepth} and Metric3Dv2~\cite{journals/corr/abs-2404-15506/metric3dv2}, even when using a lighter ResNet101~\cite{conf/cvpr/HeZRS16/resnet} backbone and a much smaller training dataset. DAC achieves superior performance across both \(360^\circ\) datasets and the fisheye dataset ScanNet++~\cite{conf/iccv/YeshwanthLND23/scannet++}. Compared to the iDisc~\cite{conf/cvpr/PiccinelliSY23/idisc} network trained with the Metric3Dv2 pipeline, DAC shows substantial improvements across all metrics on all datasets. Notably, DAC improves the next-best method by nearly 50\% in the most differentiating metric, \(\delta_1\).

In outdoor tests, DAC significantly outperforms Metric3Dv2~\cite{conf/iccv/000600CYWCS23/metric3d} and UniDepth~\cite{conf/cvpr/PiccinelliYSSLG24/unidepth}, even with much larger backbones. However, it achieves only marginal improvements over iDisc~\cite{conf/cvpr/PiccinelliSY23/idisc} under the same network configuration, with less pronounced gains compared to indoor settings. This is likely due to the limited camera pitch variance in the outdoor training data (Table~\ref{tab:dataset:overview}), reducing the ability to simulate highly distorted regions. Moreover, KITTI360~\cite{journals/pami/LiaoXG23/kitti360} LiDAR points are concentrated in less distorted areas (Fig.~\ref{fig:vis:main}), making the evaluation less distinctive.

Particularly, a notable observation is that while UniDepth~\cite{conf/cvpr/PiccinelliYSSLG24/unidepth} utilizes a network-based spherical conversion, it struggles with large FoV cameras, exposing the limitations of deep learning in extrapolated domains~\cite{conf/iclr/XuZLDKJ21/nnextrapolate}. In contrast, DAC's success underscores the effectiveness of our geometry-based training pipeline.

Additional results of DAC models using SwinL~\cite{conf/iccv/LiuL00W0LG21/swintrans} backbones are provided in Supplemental Table~\ref{tab:exp:main:2}. These models outperform their ResNet101~\cite{conf/cvpr/HeZRS16/resnet} counterparts in most cases, except on the $360^{\circ}$ datasets.

\begin{table}[hbt!]
\caption{\textbf{Impact of Key Components and Network.} We conduct the main ablation study on indoor datasets by training with HM3D~\cite{ramakrishnan2021/hm3d} and performing zero-shot testing on Pano3D-GV2~\cite{conf/cvpr/AlbanisZDGSAZD2/pano3d} and ScanNet++~\cite{conf/iccv/YeshwanthLND23/scannet++}. We compare the performance of the DAC framework with specific components removed, as well as different network architectures trained under the Metric3D~\cite{conf/iccv/000600CYWCS23/metric3d} pipeline.}
\vspace{-5mm}
\begin{center}
\begin{footnotesize}
\setlength\tabcolsep{0.1cm}
\begin{tabular}{l | l | l l l} 
\hline 
\textbf{Test Dataset} & \textbf{Methods} & $\pmb{\delta_1}$  $\uparrow$ & $\pmb{\delta_2}$  $\uparrow$ & \textbf{A.Rel}$\downarrow$ \\
\hline
\multirow{8}{*}{Pano3D-GV2 \cite{conf/cvpr/AlbanisZDGSAZD2/pano3d}}  
 & Metric3Dv2~\cite{journals/corr/abs-2404-15506/metric3dv2}  & 0.5623  & 0.8341 & 0.2479 \\
 & iDisc-cnn~\cite{conf/cvpr/PiccinelliSY23/idisc}          & 0.3026 & 0.5565 & 0.3548 \\
 & iDisc~\cite{conf/cvpr/PiccinelliSY23/idisc}              & 0.4130 & 0.6844 & 0.3043 \\ 
\cline{2-5}
 & \textbf{DAC (Ours)}                  & \textbf{0.7251} & \textbf{0.9254} & \textbf{0.1729}\\
 &  w\textbackslash o Pitch-Aware ERP   & 0.4911   & 0.7904 & 0.2422 \\
 &  w\textbackslash o Pitch Aug $10^\circ$      & 0.6912   & 0.9311 & 0.188 \\
 &   w\textbackslash o FoV Align         &  0.4075 &  0.7585 &  0.2610 \\
 &   w\textbackslash o Multi-Reso        &  0.5128 &  0.7784 &  0.2437 \\
\hline
\multirow{8}{*}{ScanNet++ \cite{conf/iccv/YeshwanthLND23/scannet++}} 
 & Metric3Dv2~\cite{journals/corr/abs-2404-15506/metric3dv2}  & 0.4569  & 0.7463 & 0.2818 \\
 & iDisc-cnn~\cite{conf/cvpr/PiccinelliSY23/idisc}          & 0.4639  & 0.7653 & 0.3045 \\
 & iDisc~\cite{conf/cvpr/PiccinelliSY23/idisc}              & 0.5301  & 0.8048 & 0.3237 \\
\cline{2-5}
 & \textbf{DAC (Ours)}                  & \textbf{0.6539}  & \textbf{0.9083} & \textbf{0.1951}\\
 &  w\textbackslash o Pitch-Aware ERP          & 0.4711   & 0.8068 & 0.2508 \\
 &  w\textbackslash o Pitch Aug $10^\circ$      & 0.6741   & 0.9066 & 0.1914 \\
 &   w\textbackslash o FoV Align         &  0.5428  &  0.8644 &  0.2200 \\
 &   w\textbackslash o Multi-Reso        &  0.5504  &  0.8464 &  0.2231 \\
\hline
\end{tabular}
\end{footnotesize}
\end{center}
\label{tab:ablation:dac}
\vspace{-4mm}
\end{table}

\begin{table}[t!]
\caption{\textbf{Impact of Train Dataset.} Models are trained separately on each training dataset and evaluated in zero-shot tests on $360^\circ$ and fisheye datasets. Due to the unique characteristics of each training dataset, their contributions and importance to generalization across different testing datasets vary.}
\vspace{-5mm}
\begin{center}
\begin{footnotesize}
\scalebox{0.9}{
\setlength\tabcolsep{0.1cm}
\begin{tabular}{l | l | l | l l} 
\hline 
\textbf{Test Datasets} & \textbf{Train Dataset} & \textbf{Methods} & $\pmb{\delta_1}$  $\uparrow$  & \textbf{A.Rel}$\downarrow$\\ 
\hline
\multirow{9}{*}{Pano3D-GV2 \cite{conf/cvpr/AlbanisZDGSAZD2/pano3d}}
 & \multirow{3}{*}{HM3D-tiny~\cite{ramakrishnan2021/hm3d}} 
 & Metric3Dv2~\cite{journals/corr/abs-2404-15506/metric3dv2}   & 0.5623  & 0.2479 \\
 && iDisc~\cite{conf/cvpr/PiccinelliSY23/idisc}               & 0.4130 & 0.3043 \\ 
 && \textbf{DAC (Ours)}                   & \textbf{0.7251} & \textbf{0.1729}\\
\cline{2-5}
 & \multirow{3}{*}{Taskonomy-tiny~\cite{conf/ijcai/ZamirSSGMS19/taskonomy}} 
 & Metric3Dv2~\cite{journals/corr/abs-2404-15506/metric3dv2}  & 0.3785  & 0.2959\\
 && iDisc~\cite{conf/cvpr/PiccinelliSY23/idisc}              & 0.3888 & 0.4076\\ 
 && \textbf{DAC (Ours)}                  & \textbf{0.6411} & \textbf{0.1972}\\
\cline{2-5}
 & \multirow{3}{*}{Hypersim~\cite{conf/iccv/RobertsRRK0PWS21/hypersim}} 
 & Metric3Dv2~\cite{journals/corr/abs-2404-15506/metric3dv2}  & 0.3085  & 0.5583\\
 && iDisc~\cite{conf/cvpr/PiccinelliSY23/idisc}              & 0.3372 & 0.3288 \\ 
 && \textbf{DAC (Ours)}                  & \textbf{0.5208} & \textbf{0.1792}\\
\hline

\hline
\multirow{9}{*}{ScanNet++ \cite{conf/iccv/YeshwanthLND23/scannet++}}
 & \multirow{3}{*}{HM3D-tiny~\cite{ramakrishnan2021/hm3d}} 
 & Metric3Dv2~\cite{journals/corr/abs-2404-15506/metric3dv2} & 0.4569 & 0.2818\\
 && iDisc~\cite{conf/cvpr/PiccinelliSY23/idisc}              & 0.5301  & 0.3237 \\
 && \textbf{DAC (Ours)}                  & \textbf{0.6539} & \textbf{0.1951}\\
 
\cline{2-5}
 & \multirow{3}{*}{Taskonomy-tiny~\cite{conf/ijcai/ZamirSSGMS19/taskonomy}} 
 & Metric3Dv2~\cite{journals/corr/abs-2404-15506/metric3dv2}  & 0.6318 & 0.2148\\
 && iDisc~\cite{conf/cvpr/PiccinelliSY23/idisc}              & 0.6743 & 0.1977\\ 
 && \textbf{DAC (Ours)}                  & \textbf{0.7981} & \textbf{0.1447}\\
\cline{2-5}
 & \multirow{3}{*}{Hypersim~\cite{conf/iccv/RobertsRRK0PWS21/hypersim}} 
 & Metric3Dv2~\cite{journals/corr/abs-2404-15506/metric3dv2}  & 0.5050  & 0.2269 \\
 && iDisc~\cite{conf/cvpr/PiccinelliSY23/idisc}              & 0.6656 & 0.2213 \\ 
 && \textbf{DAC (Ours)}                  & \textbf{0.7478} & \textbf{0.1762}\\
\hline
\end{tabular}
}
\end{footnotesize}
\end{center}
\label{tab:ablation:dataset}
\vspace{-4mm}
\end{table}

\subsection{Ablation Study}
\label{sec:ablation}

\noindent\textbf{Key Components and Network Architecture.} We evaluate the effect of the FoV Align and Multi-Reso Training components by removing each individually, while keeping the rest of the DAC framework unchanged. This ablation is conducted on the challenging HM3D-tiny~\cite{ramakrishnan2021/hm3d} indoor dataset, which includes varied camera FoVs, pitch angles, and lower-quality images from reconstructed scenes. We also test the impact of removing attention modules from iDisc~\cite{conf/cvpr/PiccinelliSY23/idisc} and compare to Metric3Dv2~\cite{journals/corr/abs-2404-15506/metric3dv2} to isolate the influence of the iDisc architecture. Both iDisc-based methods and DAC use ResNet101 backbones, while Metric3Dv2 uses Dinov2. Table~\ref{tab:ablation:dac} provides a summary; full metrics and Matterport3D~\cite{Matterport3D} results are in the Supplemental Table~\ref{tab:ablation:dac:full}.

\begin{figure}[!tb]
    \centering
    \includegraphics[width=0.47\textwidth]{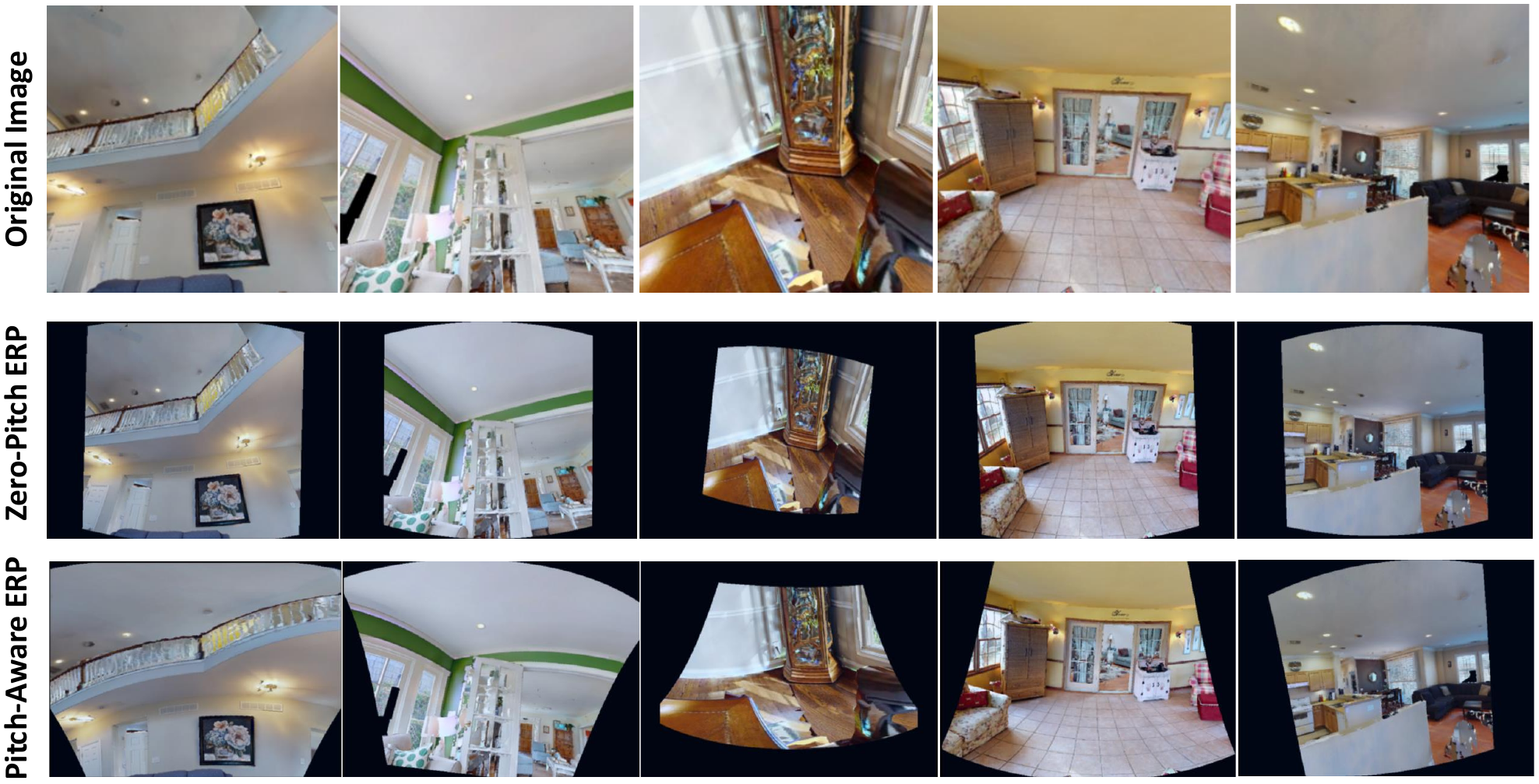}
    \caption{\textbf{Pitch-Aware Image-to-ERP Conversion.} \textit{Top}: The original images, taking HM3D~\cite{ramakrishnan2021/hm3d} samples for examples. \textit{Middle}: ERP patches converted from the original images without camera pitch awareness by setting tangent image center at latitude $\lambda_c = 0$. \textit{Bottom}: ERP patches prepared via camera pitch-aware ERP conversion, where in our convention $\lambda_c = -\text{Pitch}$.
    }
   \label{fig:pitch:aware}
\end{figure}

Table~\ref{tab:ablation:dac} highlights the pivotal role of pitch-aware ERP conversion in generalizing perspective-trained models to large FoV datasets by effectively simulating high-distortion regions uniquely observed in large FoV images (Fig.~\ref{fig:pitch:aware}). This approach turns the wide pitch angle variance in datasets like HM3D~\cite{ramakrishnan2021/hm3d} into an advantage. While additional pitch augmentation does not appear essential when the training dataset like HM3D intricately spans a large range of pitch angle. However, its effectiveness varies across datasets, as detailed in Supplemental Table~\ref{tab:ablation:dac:full}.

Results in Table~\ref{tab:ablation:dac} also show that removing FoV Align or Multi-Reso Training significantly reduces DAC performance, particularly for zero-shot generalization on \(360^\circ\) images. Compared to the iDisc network trained with the Metric3D pipeline, DAC achieves notable improvements for large FoV cameras, with attention modules in iDisc proving effective for large FoV test data. Although Metric3Dv2 uses a heavier backbone, it shows limited zero-shot generalization on large FoV images without DAC. More comprehensive results can be found both in Supplemental Table~\ref{tab:ablation:dac:full}.

\noindent\textbf{Impact of Training Dataset.} Each dataset has unique characteristics (Table~\ref{tab:dataset:overview}). To evaluate their impact on generalization to large FoV data, we trained models separately on HM3D-tiny, Taskonomy-tiny~\cite{conf/ijcai/ZamirSSGMS19/taskonomy}, and Hypersim~\cite{conf/iccv/RobertsRRK0PWS21/hypersim}, then tested them in zero-shot mode on indoor large FoV datasets. Results are summarized in Table~\ref{tab:ablation:dataset}, with full results in Supplemental Table~\ref{tab:ablation:dataset:full}.

For Pano3D-GV2~\cite{conf/cvpr/AlbanisZDGSAZD2/pano3d}, broader FoV and pitch angle coverage in HM3D training improve generalization across all methods, despite HM3D's lower quality due to rendering artifacts. For fisheye data in ScanNet++~\cite{conf/iccv/YeshwanthLND23/scannet++}, FoV diversity appears less crucial, as the single-camera Hypersim dataset, despite limited training data, outperforms HM3D, indicating that image quality plays a key role in ScanNet++ test.

Comparing individually trained models with results from mixed training in Table~\ref{tab:exp:main} shows that DAC effectively leverages the synergy of diverse datasets, significantly enhancing generalization to large FoV datasets.


\section{Conclusion}
\label{sec:conclusion}

We introduced the Depth Any Camera (DAC) framework for zero-shot metric depth estimation across diverse camera types, including perspective, fisheye, and $360^\circ$ cameras. By leveraging a highly effective pitch-aware Image-to-ERP transformation, FoV alignment, and multi-resolution training, DAC addresses the challenges posed by varying FoVs and resolution inconsistencies and enables robust generalization on large FoV datasets. Our results demonstrate that DAC significantly outperforms state-of-the-art methods and adapts seamlessly to different backbone networks. In practice, DAC ensures that every piece of previously collected 3D data remains valuable, regardless of the camera type used in new applications.

\small \bibliographystyle{ieeenat_fullname} 
\bibliography{main}



\clearpage
\setcounter{page}{1}
\maketitlesupplementary


\section{Supplemental Experiments}



\subsection{Full Zero-Shot Metric Depth Experiments}

Full experiments with a few additional experiments comparing DAC to the SoTA methods in zero-short metric depth estimation are shown in Table~\ref{tab:exp:main:2}. The additional experiments include:

\begin{itemize}
    \item \textbf{Zero-Shot to Perspective Data.} In addition to the large FoV dataset results presented in the main text, we include evaluations on two widely tested perspective datasets, NYUv2~\cite{Silberman:ECCV12:nyuv2} and KITTI~\cite{journals/ijrr/GeigerLSU13/kitti}, to demonstrate that our method can also achieve zero-shot generalization on standard perspective datasets. Notably, DAC outperforms iDisc~\cite{conf/cvpr/PiccinelliSY23/idisc} trained with the Metric3Dv2~\cite{journals/corr/abs-2404-15506/metric3dv2} pipeline, which we attribute to DAC's ability to leverage the synergy of diverse data with varying FoVs and pitch coverage. The remaining gap compared to the state-of-the-art is likely due to the significantly smaller training dataset and the smaller SwinL~\cite{conf/iccv/LiuL00W0LG21/swintrans} backbone used in DAC compared to the larger ViT-L~\cite{conf/iclr/DosovitskiyB0WZ21/vit} backbones adopted by other methods.

    \item \textbf{DAC with SwinL~\cite{conf/iccv/LiuL00W0LG21/swintrans} Backbone.} We also update our DAC model and iDisc model with a larger backbone, Swin-L~\cite{conf/iccv/LiuL00W0LG21/swintrans}, to further showcase the performance of our approach when scaling to larger models. Note that the Swin-L backbone remains smaller than the Dinov2-ViT-L~\cite{journals/tmlr/OquabDMVSKFHMEA24/dinov2} backbone used in Metric3Dv2~\cite{journals/corr/abs-2404-15506/metric3dv2}, and as well the ViT-L~\cite{conf/iclr/DosovitskiyB0WZ21/vit} backbone applied in UniDepth~\cite{conf/cvpr/PiccinelliYSSLG24/unidepth}. As observed, although Swin-L-based DAC models lead to significant improvements on generalization to NYU and KITTI360 datasets, their improvements on Scannet++ and KITTI datasets are marginal, and they under perform Resnet101 counterparts on $360^\circ$ datasets. We interpreter the reason is that transformer backbones are designed for scale-invariance reasoning rather than for the scale-equivariance inference required in 3D tasks. More adapted design of transformer architectures are demanding for further push the upper bound of training of foundation depth models.


\end{itemize}

\begin{table*}[t!]
\caption{
\textbf{Zero-Shot Metric Depth Evaluation on $\mathbf{360^\circ}$, Fisheye, and Perspective Datasets.} This table compares DAC with leading state-of-the-art metric depth models across metric depth benchmarks, upon Resnet101~\cite{conf/cvpr/HeZRS16/resnet} and SwinL~\cite{conf/iccv/LiuL00W0LG21/swintrans} backbones.
}
\vspace{-5mm}
\begin{center}
\begin{footnotesize}
\setlength\tabcolsep{0.1cm}
\begin{tabular}{l | l | l | l | l l l l  l  l } 
\hline 
\textbf{Test Dataset} & \textbf{Methods} & \textbf{Train Dataset} & \textbf{Backbone} & $\pmb{\delta_1}$  $\uparrow$ & $\pmb{\delta_2}$  $\uparrow$ & $\pmb{\delta_3}$  $\uparrow$  & \textbf{Abs Rel}$\downarrow$ & \textbf{RMSE}$\downarrow$ & \textbf{log10}$\downarrow$  \\
\hline
\multirow{7}{*}{Matterport3D \cite{Matterport3D}} 
 & UniDepth~\cite{conf/cvpr/PiccinelliYSSLG24/unidepth}         & Mix 3M      & ViT-L~\cite{conf/iclr/DosovitskiyB0WZ21/vit}     & 0.2576  & 0.5114 & 0.7091 & 0.7648 & 1.3827 & 0.2208\\
 & Metric3Dv2~\cite{journals/corr/abs-2404-15506/metric3dv2}    & Mix 16M     & Dinov2-ViT-L~\cite{journals/tmlr/OquabDMVSKFHMEA24/dinov2}    & 0.4381  & 0.7311 & 0.8735 & 0.2924 & 0.8842 & 0.1546\\
 & Metric3Dv2~\cite{journals/corr/abs-2404-15506/metric3dv2}    & Indoor 670K & Dinov2-ViT-L~\cite{journals/tmlr/OquabDMVSKFHMEA24/dinov2}    & 0.4287  & 0.7854 & 0.9333 & 0.2788 & 0.8961 & 0.1352\\
 & iDisc~\cite{conf/cvpr/PiccinelliSY23/idisc}                  & Indoor 670K & Resnet101~\cite{conf/cvpr/HeZRS16/resnet} & 0.5287 & 0.8260 & 0.9398 & 0.2757 & 0.7771 & 0.1147\\ 
 & iDisc~\cite{conf/cvpr/PiccinelliSY23/idisc}                  & Indoor 670K & SwinL~\cite{conf/iccv/LiuL00W0LG21/swintrans} & 0.5865 & 0.8722 & 0.9599 & 0.2272 & 0.6612 & 0.1021\\ 
 & \textbf{DAC (Ours)}                       & Indoor 670K & Resnet101~\cite{conf/cvpr/HeZRS16/resnet} & \textbf{0.7727} & \textbf{0.9562} & \textbf{0.9822} & \textbf{0.156} & \textbf{0.6185} & \textbf{0.0707}\\
 & \textbf{DAC (Ours)}                       & Indoor 670K & SwinL~\cite{conf/iccv/LiuL00W0LG21/swintrans} & 0.7231 & 0.949 & 0.9866 & 0.1789 & 0.5911 & 0.0741\\

\hline
\multirow{7}{*}{Pano3D-GV2 \cite{conf/cvpr/AlbanisZDGSAZD2/pano3d}}  
 & UniDepth~\cite{conf/cvpr/PiccinelliYSSLG24/unidepth}         & Mix 3M      & ViT-L~\cite{conf/iclr/DosovitskiyB0WZ21/vit}     & 0.2469  & 0.4977 & 0.7084 & 0.7892 & 1.2681 & 0.2231\\
 & Metric3Dv2~\cite{journals/corr/abs-2404-15506/metric3dv2}    & 16M         & Dinov2-ViT-L~\cite{journals/tmlr/OquabDMVSKFHMEA24/dinov2}    & 0.4040  & 0.6929 & 0.8499 & 0.3070 & 0.8549 & 0.1664\\
 & Metric3Dv2~\cite{journals/corr/abs-2404-15506/metric3dv2}    & Indoor 670K & Dinov2-ViT-L~\cite{journals/tmlr/OquabDMVSKFHMEA24/dinov2}    & 0.5060  & 0.8176 & 0.9360 & 0.2608 & 0.7248 & 0.1201\\
 & iDisc~\cite{conf/cvpr/PiccinelliSY23/idisc}                  & Indoor 670K & Resnet101~\cite{conf/cvpr/HeZRS16/resnet} & 0.5629 & 0.8222 & 0.9332 & 0.2657 & 0.6446 & 0.1122\\ 
 & iDisc~\cite{conf/cvpr/PiccinelliSY23/idisc}                  & Indoor 670K & SwinL~\cite{conf/iccv/LiuL00W0LG21/swintrans} & 0.6022 & 0.8528 & 0.9447 & 0.2272 & 0.5680 & 0.1035\\ 
 & \textbf{DAC (Ours)}                      & Indoor 670K & Resnet101~\cite{conf/cvpr/HeZRS16/resnet} & \textbf{0.8115} & \textbf{0.9549} & \textbf{0.9860} & \textbf{0.1387} & \textbf{0.4780} & \textbf{0.0623}\\
 & \textbf{DAC (Ours)}                      & Indoor 670K & SwinL~\cite{conf/iccv/LiuL00W0LG21/swintrans} & 0.7287 & 0.9307 & 0.9793 & 0.1836 & 0.4833 & 0.077\\

\hline
\multirow{7}{*}{ScanNet++ \cite{conf/iccv/YeshwanthLND23/scannet++}} 
 & UniDepth~\cite{conf/cvpr/PiccinelliYSSLG24/unidepth}         & Mix 3M      & ViT-L~\cite{conf/iclr/DosovitskiyB0WZ21/vit}     & 0.3638  & 0.6461 & 0.8358 & 0.4971 & 1.1659 & 0.1648\\
 & Metric3Dv2~\cite{journals/corr/abs-2404-15506/metric3dv2}    & Mix 16M     & Dinov2-ViT-L~\cite{journals/tmlr/OquabDMVSKFHMEA24/dinov2}    & 0.5360  & 0.8218 & 0.9350 & 0.2229 & 0.8950 & 0.1177\\
 & Metric3Dv2~\cite{journals/corr/abs-2404-15506/metric3dv2}    & Indoor 670K & Dinov2-ViT-L~\cite{journals/tmlr/OquabDMVSKFHMEA24/dinov2}    & 0.6489  & 0.8920 & 0.9558 & 0.1915 & 0.9779 & 0.0938\\
 & iDisc~\cite{conf/cvpr/PiccinelliSY23/idisc}                  & Indoor 670K & Resnet101~\cite{conf/cvpr/HeZRS16/resnet} & 0.6150  & 0.8780 & 0.9617 & 0.2712 & 0.4835 & 0.0972\\
 & iDisc~\cite{conf/cvpr/PiccinelliSY23/idisc}                  & Indoor 670K & Swinl~\cite{conf/iccv/LiuL00W0LG21/swintrans} & 0.7746  & 0.9439 & 0.9862 & 0.1741 & 0.3634 & 0.0680\\
 & \textbf{DAC (Ours)}                      & Indoor 670K & Resnet101~\cite{conf/cvpr/HeZRS16/resnet} & 0.8517  & 0.9693 & 0.9922 & 0.1323 & 0.3086 & 0.0532\\
 & \textbf{DAC (Ours)}                      & Indoor 670K & SwinL~\cite{conf/iccv/LiuL00W0LG21/swintrans} & \textbf{0.8544}  & \textbf{0.9776} & \textbf{0.9939} & \textbf{0.1282} & \textbf{0.2866} & \textbf{0.0518}\\
\hline
\multirow{7}{*}{KITTI360 \cite{journals/pami/LiaoXG23/kitti360}} 
 & UniDepth~\cite{conf/cvpr/PiccinelliYSSLG24/unidepth}         & Mix 3M       & ViT-L~\cite{conf/iclr/DosovitskiyB0WZ21/vit}     & 0.4810  & 0.8397 & 0.9406 & 0.2939 & 6.5642 & 0.1221\\
 & Metric3Dv2~\cite{journals/corr/abs-2404-15506/metric3dv2}    & Mix 16M      & Dinov2-ViT-L~\cite{journals/tmlr/OquabDMVSKFHMEA24/dinov2}    & 0.7159  & 0.9323 & 0.9771 & 0.1997 & 4.5769 & 0.0811\\
 & Metric3Dv2~\cite{journals/corr/abs-2404-15506/metric3dv2}    & Outdoor 130K & Dinov2-ViT-L~\cite{journals/tmlr/OquabDMVSKFHMEA24/dinov2}    & 0.7675  & 0.9370 & 0.9756 & 0.1521 & 4.6610 & 0.0723\\
 & iDisc~\cite{conf/cvpr/PiccinelliSY23/idisc}                  & Outdoor 130K & Resnet101~\cite{conf/cvpr/HeZRS16/resnet} & 0.7833  & 0.9384 & 0.9753 & 0.1598 & 4.9122 & 0.0704\\
 & iDisc~\cite{conf/cvpr/PiccinelliSY23/idisc}                  & Outdoor 130K & SwinL~\cite{conf/iccv/LiuL00W0LG21/swintrans} & 0.8165  & 0.9533 & 0.9829 & 0.1500 & 4.2549 & 0.0620\\
 & \textbf{DAC (Ours)}                      & Outdoor 130K & Resnet101~\cite{conf/cvpr/HeZRS16/resnet} & 0.7858  & 0.9388 & 0.9775 & 0.1559 & 4.3614 & 0.0684\\
 & \textbf{DAC (Ours)}                      & Outdoor 130K & SwinL~\cite{conf/iccv/LiuL00W0LG21/swintrans} & \textbf{0.8222}  & \textbf{0.9571} & \textbf{0.9845} & \textbf{0.1487} & \textbf{3.7510} & \textbf{0.0607}\\
\hline

\multirow{7}{*}{NYUv2 \cite{Silberman:ECCV12:nyuv2}} 
 & UniDepth~\cite{conf/cvpr/PiccinelliYSSLG24/unidepth}         & Mix 3M  & ViT-L~\cite{conf/iclr/DosovitskiyB0WZ21/vit}    & \textbf{0.9875}  & \textbf{0.9982} & \textbf{0.9995} & \textbf{0.052} & \textbf{0.1936} & \textbf{0.0223}\\
 & Metric3Dv2~\cite{journals/corr/abs-2404-15506/metric3dv2}    & Mix 16M  & Dinov2-ViT-L \cite{journals/tmlr/OquabDMVSKFHMEA24/dinov2}   & 0.9718  & 0.9929 & 0.9971 & 0.0666 & 0.2621 & 0.0290\\
 & Metric3Dv2~\cite{journals/corr/abs-2404-15506/metric3dv2}    & Indoor 670K & Dinov2-ViT-L \cite{journals/tmlr/OquabDMVSKFHMEA24/dinov2} & 0.9422  & 0.9885 & 0.9966 & 0.0936 & 0.3359 & 0.0388\\
 & iDisc~\cite{conf/cvpr/PiccinelliSY23/idisc}                  & Indoor 670K & Resnet101~\cite{conf/cvpr/HeZRS16/resnet} & 0.691  & 0.9028 & 0.9675 & 0.1755 & 0.6193 & 0.0838\\
 & iDisc~\cite{conf/cvpr/PiccinelliSY23/idisc}                  & Indoor 670K & SwinL \cite{conf/iccv/LiuL00W0LG21/swintrans} & 0.8319  & 0.9629 & 0.9891 & 0.1239 & 0.4690 & 0.0571\\
 & \textbf{DAC (Ours)}                                          & Indoor 670K & Resnet101~\cite{conf/cvpr/HeZRS16/resnet} & 0.719  & 0.9324 & 0.985 & 0.1641 & 0.6189 & 0.0755\\
 & \textbf{DAC (Ours)}                                          & Indoor 670K & SwinL~\cite{conf/iccv/LiuL00W0LG21/swintrans} & 0.8673  & 0.975 & 0.9921 & 0.1187 & 0.4471 & 0.0511\\
\hline

\multirow{7}{*}{KITTI \cite{journals/ijrr/GeigerLSU13/kitti}} 
 & UniDepth~\cite{conf/cvpr/PiccinelliYSSLG24/unidepth}         & Mix 3M    & ViT-L \cite{conf/iclr/DosovitskiyB0WZ21/vit}   & 0.9643  & \textbf{0.9973} & \textbf{0.9993} & 0.1159 & 2.7881 & 0.047\\
 & Metric3Dv2~\cite{journals/corr/abs-2404-15506/metric3dv2}    & Mix 16M  & Dinov2-ViT-L \cite{journals/tmlr/OquabDMVSKFHMEA24/dinov2}    & \textbf{0.9742}  & 0.9954 & 0.9987 & \textbf{0.0534} & \textbf{2.4932} & \textbf{0.0234}\\
 & Metric3Dv2~\cite{journals/corr/abs-2404-15506/metric3dv2}    & Outdoor 130K & Dinov2-ViT-L \cite{journals/tmlr/OquabDMVSKFHMEA24/dinov2} & 0.9488  & 0.9918 & 0.9975 & 0.0848 & 3.1426 & 0.0375\\
 & iDisc~\cite{conf/cvpr/PiccinelliSY23/idisc}                  & Outdoor 130K & Resnet101~\cite{conf/cvpr/HeZRS16/resnet} & 0.8503  & 0.9626 & 0.9897 & 0.1277 & 4.5347 & 0.0528\\
 & iDisc~\cite{conf/cvpr/PiccinelliSY23/idisc}                  & Outdoor 130K & SwinL \cite{conf/iccv/LiuL00W0LG21/swintrans} & 0.8382  & 0.9682 & 0.993 & 0.1439 & 4.5267 & 0.0575\\

 & \textbf{DAC (Ours)}                                          & Outdoor 130K & Resnet101~\cite{conf/cvpr/HeZRS16/resnet} & 0.8767  & 0.9744 & 0.9934 & 0.1155 & 4.3877 & 0.0488\\
 & \textbf{DAC (Ours)}                                          & Outdoor 130K & SwinL \cite{conf/iccv/LiuL00W0LG21/swintrans} & 0.8912  & 0.9785 & 0.9947 & 0.1058 & 4.1699 & 0.0435\\
\hline

\end{tabular}
\end{footnotesize}
\end{center}
\label{tab:exp:main:2}
\end{table*}


\subsection{Full Modular Ablation Study}

Table~\ref{tab:ablation:dac:full} presents the complete experimental results for the ablation study of DAC's key components: \textbf{pitch-aware ERP conversion} and \textbf{pitch augmentation}, \textbf{FoV-Align}, and \textbf{Multi-Reso Training}.  It also includes comparisons to alternative network architectures and training frameworks. All the methods presented in this table are training on HM3D-tiny~\cite{ramakrishnan2021/hm3d} including about 300K samples. iDisc~\cite{conf/cvpr/PiccinelliSY23/idisc}-based and DAC models are all based on Resnet101~\cite{conf/cvpr/HeZRS16/resnet} backbone, and trained with 40K iterations with batch size 48. While Metric3Dv2~\cite{journals/corr/abs-2404-15506/metric3dv2} model is based on its original Dinov2-ViT-L~\cite{journals/tmlr/OquabDMVSKFHMEA24/dinov2} backbone, trained on the same dataset with 120K iterations and batch size 48.

The \textbf{pitch-aware ERP conversion} and \textbf{ERP-space pitch augmentation} ablations, highlight the effectiveness of our core Image-to-ERP conversion in enabling the DAC framework. 
As shown in Table~\ref{tab:ablation:dac:full}, pitch-aware ERP conversion plays a pivotal role in generalizing perspective-trained models to large FoV datasets. This capability stems from projecting input images to different latitude regions of the ERP space—areas typically visible only in large FoV data—illustrated in Fig.~\ref{fig:pitch:aware}. By leveraging this approach, the wide pitch angle variance in datasets like HM3D~\cite{ramakrishnan2021/hm3d} becomes a strength rather than a challenge. 

Note that the camera orientations wrt. the world coordinates can be either provided by the dataset~\cite{ramakrishnan2021/hm3d,conf/ijcai/ZamirSSGMS19/taskonomy,conf/iccv/RobertsRRK0PWS21/hypersim}, or estimated from tradition geometry~\cite{schoenberger2016sfm} or recent deep learning models~\cite{DBLP:conf/cvpr/JinZHWBSF23/perspectivefield}. Since our training process is usually integrated with ERP space geometric augmentations, our framework do not require the camera pose estimation very accurate for the purpose of depth estimation.

Additionally, ERP-space pitch augmentation provides marginal improvements for $360^\circ$ datasets and minimal gains for Scannet++ fisheye data, likely because HM3D-tiny already includes a sufficiently broad pitch span.


\begin{table*}[t!]
\caption{\textbf{Impact of Key Components and Network.} We conduct the main ablation study on indoor datasets by training with HM3D~\cite{ramakrishnan2021/hm3d} and performing zero-shot testing on Pano3D-GV2~\cite{conf/cvpr/AlbanisZDGSAZD2/pano3d} and ScanNet++\cite{conf/iccv/YeshwanthLND23/scannet++}. We compare the performance of the DAC framework with specific components removed, as well as different network architectures trained under the Metric3D\cite{conf/iccv/000600CYWCS23/metric3d} pipeline. Four key components of our DAC framework are included in the ablation study.}
\begin{center}
\begin{footnotesize}
\begin{tabular}{l | l | l l l l l l} 
\hline 
\textbf{Test Datasets} & \textbf{Methods} & $\pmb{\delta_1}$  $\uparrow$ & $\pmb{\delta_2}$  $\uparrow$ & $\pmb{\delta_3}$  $\uparrow$  & \textbf{Abs Rel}$\downarrow$ & \textbf{RMSE}$\downarrow$ & \textbf{log10}$\downarrow$  \\
\hline
\hline
\multirow{8}{*}{Matterport3D \cite{Matterport3D}} 
 & Metric3Dv2~\cite{journals/corr/abs-2404-15506/metric3dv2}  & 0.4879  & 0.8196 & 0.9443 & 0.2631 & 0.8556 & 0.1214\\
 & iDisc-cnn~\cite{conf/cvpr/PiccinelliSY23/idisc}          & 0.3574 & 0.6355 & 0.8051 & 0.3202 & 1.3369 & 0.1854\\
 & iDisc~\cite{conf/cvpr/PiccinelliSY23/idisc}              & 0.4303 & 0.7325 & 0.8777 & 0.3109 & 1.1876 & 0.1508\\ 
  & \textbf{DAC (Ours)}                          & \textbf{0.728} & \textbf{0.9372} & \textbf{0.9761} & \textbf{0.1699} & \textbf{0.718} & \textbf{0.0774}\\
  &  w\textbackslash o Pitch-Aware ERP          & 0.5394   & 0.8358 & 0.9442 & 0.2222 & 0.8383 & 0.1134\\
  &  w\textbackslash o Pitch Aug $10^\circ$     & 0.7152   & 0.9379 & 0.9797 & 0.1816 & 0.7134 & 0.0789\\
  &  w\textbackslash o FoV Align                 & 0.4494  & 0.7962 & 0.9206 & 0.2446 & 1.0383 & 0.1331\\
  &  w\textbackslash o Multi-Reso                & 0.5670   & 0.8476 & 0.9343 & 0.2219 & 0.9658 & 0.1132\\
  
\hline
\multirow{8}{*}{Pano3D-GV2 \cite{conf/cvpr/AlbanisZDGSAZD2/pano3d}}  
 & Metric3Dv2~\cite{journals/corr/abs-2404-15506/metric3dv2}  & 0.5623  & 0.8341 & 0.9396 & 0.2479 & 0.7332 & 0.1113\\
 & iDisc-cnn~\cite{conf/cvpr/PiccinelliSY23/idisc}          & 0.3026 & 0.5565 & 0.7337 & 0.3548 & 1.2307 & 0.2118\\
 & iDisc~\cite{conf/cvpr/PiccinelliSY23/idisc}              & 0.413 & 0.6844 & 0.8397 & 0.3043 & 1.0649 & 0.162\\ 
 &  \textbf{DAC (Ours)}                  & \textbf{0.7251} & \textbf{0.9254} & \textbf{0.9747} & \textbf{0.1729} & \textbf{0.6015} & \textbf{0.0786}\\
 &  w\textbackslash o Pitch-Aware ERP          & 0.4911   & 0.7904 & 0.9193 & 0.2422 & 0.7521 & 0.1262\\
 &  w\textbackslash o Pitch Aug $10^\circ$      & 0.6912   & 0.9311 & 0.977 & 0.188 & 0.5966 & 0.0819\\
 &  w\textbackslash o FoV Align                 & 0.4075 & 0.7585 & 0.9085 & 0.261 & 0.9148 & 0.1415\\
 &  w\textbackslash o Multi-Reso                & 0.5128 & 0.7784 & 0.8977 & 0.2437 & 0.8867 & 0.1298\\
\hline
\multirow{8}{*}{ScanNet++ \cite{conf/iccv/YeshwanthLND23/scannet++}} 
 & Metric3Dv2~\cite{journals/corr/abs-2404-15506/metric3dv2}  & 0.3865  & 0.6730 & 0.8229 & 0.3129 & 1.3277 & 0.1705\\
 & iDisc-cnn~\cite{conf/cvpr/PiccinelliSY23/idisc}          & 0.4639  & 0.7653 & 0.8965 & 0.3045 & 1.3116 & 0.1395\\
 & iDisc~\cite{conf/cvpr/PiccinelliSY23/idisc}              & 0.5301  & 0.8048 & 0.9165 & 0.3237 & 1.552 & 0.1251\\
 & \textbf{DAC (Ours)}                          & \textbf{0.6539}  & \textbf{0.9083} & \textbf{0.9722} & \textbf{0.1951} & \textbf{0.5926} & \textbf{0.089}\\
 &  w\textbackslash o Pitch-Aware ERP          & 0.4711   & 0.8068 & 0.9282 & 0.2508 & 0.7925 & 0.127\\
 &  w\textbackslash o Pitch Aug $10^\circ$      & 0.6741   & 0.9066 & 0.9701 & 0.1914 & 0.5966 & 0.0861\\
 &  w\textbackslash o FoV Align                 & 0.5428  & 0.8644 & 0.9544 & 0.22 & 0.71 & 0.1091\\
 &  w\textbackslash o Multi-Reso                & 0.5504  & 0.8464 & 0.942 & 0.2231 & 0.7435 & 0.1116\\

\hline

\end{tabular}
\end{footnotesize}
\end{center}
\label{tab:ablation:dac:full}
\end{table*}


\subsection{Full Ablation Study on Training Dataset}

In Table~\ref{tab:ablation:dataset:full}, we show the full ablation study on the impact of different datasets. Different training dataset, due to its different span in camera FoVs, pitch angles, image quality, etc., contribute differently on different testing data. Our DAC framework can leverage the synergy between very diverse datasets to significantly boost the overall performance to all the testing datasets.

In addition to the main content summarized in the paper,we include an ablation study on the impact of \textbf{pitch-aware ERP conversion} and \textbf{ERP-space pitch augmentation} to evaluate their effectiveness across different training datasets.

The results indicate that pitch-aware ERP conversion is crucial for DAC's generalization across almost all configurations of training and testing datasets. This remains true even when the training dataset has a limited range of camera pitch angles, such as Taskonomy~\cite{conf/ijcai/ZamirSSGMS19/taskonomy}. Moreover, its impact becomes more pronounced as the diversity of pitch angles in the training dataset increases. In contrast, ERP-space pitch augmentation proves significant primarily when the original training dataset lacks diversity in pitch angles. However, its contribution diminishes when the training data already encompass a wide range of pitch angles.


\begin{table*}[hbt!]
\caption{\textbf{Ablation Study of training datasets.} Models are trained separately on each training dataset and evaluated in zero-shot tests on $360^\circ$ and fisheye datasets. In addition, the ablation study on the impact of pitch-aware ERP conversion and ERP-space pitch augmentation are included to further analysis their contribution under different training distributions.
}
\begin{center}
\begin{footnotesize}
\setlength\tabcolsep{0.1cm}
\begin{tabular}{l | l | l | l l l l  l  l } 
\hline 
\textbf{Test Datasets} & \textbf{Train Dataset} & \textbf{Methods} & $\pmb{\delta_1}$  $\uparrow$ & $\pmb{\delta_2}$  $\uparrow$ & $\pmb{\delta_3}$  $\uparrow$  & \textbf{Abs Rel}$\downarrow$ & \textbf{RMSE}$\downarrow$ & \textbf{log10}$\downarrow$  \\ 
\hline
\hline
\multirow{15}{*}{Matterport3D \cite{Matterport3D}} 
 & \multirow{5}{*}{HM3D-tiny~\cite{ramakrishnan2021/hm3d} 310K} 
 & Metric3Dv2~\cite{journals/corr/abs-2404-15506/metric3dv2}   & 0.4879  & 0.8196 & 0.9443 & 0.2631 & 0.8556 & 0.1214\\
 && iDisc~\cite{conf/cvpr/PiccinelliSY23/idisc}              & 0.4303 & 0.7325 & 0.8777 & 0.3109 & 1.1876 & 0.1508\\ 
 && \textbf{DAC (Ours)}                                      & \textbf{0.728} & \textbf{0.9372} & \textbf{0.9761} & \textbf{0.1699} & \textbf{0.718} & \textbf{0.0774}\\
 &&  w\textbackslash o Pitch-Aware ERP          & 0.5394   & 0.8358 & 0.9442 & 0.2222 & 0.8383 & 0.1134\\
&&  w\textbackslash o Pitch Aug $10^\circ$     & 0.7152   & 0.9379 & 0.9797 & 0.1816 & 0.7134 & 0.0789\\
\cline{2-9}
 & \multirow{5}{*}{Taskonomy-tiny~\cite{conf/ijcai/ZamirSSGMS19/taskonomy} 300K} 
 & Metric3Dv2~\cite{journals/corr/abs-2404-15506/metric3dv2}   & 0.3244  & 0.6652 & 0.8958 & 0.3145 & 1.0727 & 0.1711\\
 && iDisc~\cite{conf/cvpr/PiccinelliSY23/idisc}              & 0.3662 & 0.6538 & 0.8205 & 0.4186 & 2.3299 & 0.1787\\ 
 && \textbf{DAC (Ours)}                                      & \textbf{0.5363} & \textbf{0.8537} & \textbf{0.9371} & \textbf{0.232} & \textbf{0.8194} & \textbf{0.115}\\
 &&  w\textbackslash o Pitch-Aware ERP          & 0.4018   & 0.7576 & 0.894 & 0.2722 & 0.9377 & 0.1471\\
&&  w\textbackslash o Pitch Aug $10^\circ$     & 0.4244   & 0.7633 & 0.9019 & 0.2689 & 0.9199 & 0.1428\\
\cline{2-9}
 & \multirow{5}{*}{Hypersim~\cite{conf/iccv/RobertsRRK0PWS21/hypersim} 60k} 
 & Metric3Dv2~\cite{journals/corr/abs-2404-15506/metric3dv2}   &  0.3740  & 0.6746 & 0.8450 & 0.5082 & 1.0822 & 0.1637\\
 && iDisc~\cite{conf/cvpr/PiccinelliSY23/idisc}              & 0.3624 & 0.6792 & 0.8757 & 0.315  & 1.0425 & 0.1638\\ 
 && \textbf{DAC (Ours)}                                      & \textbf{0.4491} & \textbf{0.8066} & \textbf{0.9438} & \textbf{0.2659} & \textbf{0.8574} & \textbf{0.1271}\\
 &&  w\textbackslash o Pitch-Aware ERP          & 0.4098   & 0.7526 & 0.9129 & 0.2772 & 0.9437 & 0.1431\\
&&  w\textbackslash o Pitch Aug $10^\circ$     & 0.4577   & 0.834 & 0.9524 & 0.2513 & 0.8926 & 0.1206\\
\hline

\hline
\multirow{15}{*}{Pano3D-GV2 \cite{conf/cvpr/AlbanisZDGSAZD2/pano3d}}
 & \multirow{5}{*}{HM3D-tiny~\cite{ramakrishnan2021/hm3d} 310K} 
 & Metric3Dv2~\cite{journals/corr/abs-2404-15506/metric3dv2}   & 0.5623  & 0.8341 & 0.9396 & 0.2479 & 0.7332 & 0.1113\\
 && iDisc~\cite{conf/cvpr/PiccinelliSY23/idisc}              & 0.413 & 0.6844 & 0.8397 & 0.3043 & 1.0649 & 0.162\\ 
 && \textbf{DAC (Ours)}                                      & \textbf{0.7251} & \textbf{0.9254} & \textbf{0.9747} & \textbf{0.1729} & \textbf{0.6015} & \textbf{0.0786}\\
&&  w\textbackslash o Pitch-Aware ERP          & 0.4911   & 0.7904 & 0.9193 & 0.2422 & 0.7521 & 0.1262\\
&&  w\textbackslash o Pitch Aug $10^\circ$      & 0.6912   & 0.9311 & 0.977 & 0.188 & 0.5966 & 0.0819\\
\cline{2-9}
 & \multirow{5}{*}{Taskonomy-tiny~\cite{conf/ijcai/ZamirSSGMS19/taskonomy} 300K} 
 & Metric3Dv2~\cite{journals/corr/abs-2404-15506/metric3dv2}  & 0.3785  & 0.7489 & 0.9062 & 0.2959 & 0.8945 & 0.1550\\
 && iDisc~\cite{conf/cvpr/PiccinelliSY23/idisc}              & 0.3888 & 0.6816 & 0.8349 & 0.4076 & 2.1877 & 0.1683\\ 
 && \textbf{DAC (Ours)}                                      & \textbf{0.6411} & \textbf{0.8719} & \textbf{0.9452} & \textbf{0.1972} & \textbf{0.6148} & \textbf{0.0982}\\
 &&  w\textbackslash o Pitch-Aware ERP          & 0.4828   & 0.7882 & 0.9026 & 0.2465 & 0.7345 & 0.1323\\
&&  w\textbackslash o Pitch Aug $10^\circ$     & 0.4954   & 0.7947 & 0.9077 & 0.2411 & 0.7197 & 0.1289\\
\cline{2-9}
 & \multirow{5}{*}{Hypersim~\cite{conf/iccv/RobertsRRK0PWS21/hypersim} 60k} 
 & Metric3Dv2~\cite{journals/corr/abs-2404-15506/metric3dv2}  & 0.3085  & 0.6382 & 0.8147 & 0.5583 & 1.1762 & 0.1887\\
 && iDisc~\cite{conf/cvpr/PiccinelliSY23/idisc}              & 0.3372 & 0.6473 & 0.831 & 0.3288 & 0.9098 & 0.177\\ 
 && \textbf{DAC (Ours)}                                      & \textbf{0.5208} & \textbf{0.8295} & \textbf{0.9424} & \textbf{0.1792} & \textbf{0.6873} & \textbf{0.1158}\\
 &&  w\textbackslash o Pitch-Aware ERP          & 0.4486   & 0.7655 & 0.9025 & 0.2707 & 0.7823 & 0.1385\\
&&  w\textbackslash o Pitch Aug $10^\circ$     & 0.5293   & 0.8525 & 0.9504 & 0.2344 & 0.7212 & 0.1123\\
\hline

\hline
\multirow{15}{*}{ScanNet++ \cite{conf/iccv/YeshwanthLND23/scannet++}}
 & \multirow{5}{*}{HM3D-tiny~\cite{ramakrishnan2021/hm3d} 310K} 
 & Metric3Dv2~\cite{journals/corr/abs-2404-15506/metric3dv2}   & 0.3799  & 0.6310 & 0.7801 & 0.6090 & 1.0490 & 0.1899\\
 && iDisc~\cite{conf/cvpr/PiccinelliSY23/idisc}              & 0.5301  & 0.8048 & 0.9165 & 0.3237 & 1.552 & 0.1251\\
 && \textbf{DAC (Ours)}                                      & \textbf{0.6539}  & \textbf{0.9083} & \textbf{0.9722} & \textbf{0.1951} & \textbf{0.5926} & \textbf{0.089}\\
 &&  w\textbackslash o Pitch-Aware ERP          & 0.4711   & 0.8068 & 0.9282 & 0.2508 & 0.7925 & 0.127\\
&&  w\textbackslash o Pitch Aug $10^\circ$      & 0.6741   & 0.9066 & 0.9701 & 0.1914 & 0.5966 & 0.0861\\
\cline{2-9}
 & \multirow{5}{*}{Taskonomy-tiny~\cite{conf/ijcai/ZamirSSGMS19/taskonomy} 300K} 
 & Metric3Dv2~\cite{journals/corr/abs-2404-15506/metric3dv2} & 0.6421 & 0.8377 & 0.9285 & 0.3840 & 2.2102 & 0.1075\\
 && iDisc~\cite{conf/cvpr/PiccinelliSY23/idisc}              & 0.6743 & 0.9179 & 0.9809 & 0.1977 & 0.5235 & 0.083\\ 
 && \textbf{DAC (Ours)}                                      & \textbf{0.7981} & \textbf{0.9666} & \textbf{0.9898} & \textbf{0.1447} & \textbf{0.3556} & \textbf{0.0637}\\
 &&  w\textbackslash o Pitch-Aware ERP          & 0.7642   & 0.9561 & 0.9879 & 0.1542 & 0.3881 & 0.0705\\
&&  w\textbackslash o Pitch Aug $10^\circ$     & 0.7673   & 0.9534 & 0.9892 & 0.1516 & 0.3861 & 0.0694\\
\cline{2-9}
 & \multirow{5}{*}{Hypersim~\cite{conf/iccv/RobertsRRK0PWS21/hypersim} 60k} 
 & Metric3Dv2~\cite{journals/corr/abs-2404-15506/metric3dv2}  & 0.5684  & 0.8149 & 0.9173 & 0.3364 & 0.5289 & 0.1192\\
 && iDisc~\cite{conf/cvpr/PiccinelliSY23/idisc}              & 0.6656 & 0.9004 & 0.9701 & 0.2213 & 0.5471 & 0.0872\\ 
 && \textbf{DAC (Ours)}                                      & \textbf{0.7478} & \textbf{0.9483} & \textbf{0.9871} & \textbf{0.1762} & \textbf{0.4124} & \textbf{0.0729}\\
 &&  w\textbackslash o Pitch-Aware ERP          & 0.7238   & 0.9236 & 0.9801 & 0.1959 & 0.4375 & 0.0778\\
&&  w\textbackslash o Pitch Aug $10^\circ$     & 0.7439   & 0.9396 & 0.9844 & 0.1846 & 0.4106 & 0.0732\\
\hline

\end{tabular}
\end{footnotesize}
\end{center}
\label{tab:ablation:dataset:full}
\end{table*}


\subsection{Zero-Shot Test of Perspective Depth Model on Distorted Images}

As shown in Table~\ref{tab:metric:depth:zero-shot:distortion}, we evaluate Metric3D~\cite{journals/corr/abs-2404-15506/metric3dv2} on different representations of KITTI360's fisheye images including raw fisheye, the ERP conversion of fisheye, undistorted fisheye with three different FoVs. 
The evaluation results align with the visual examples in Figure~\ref{fig:issue:fisheye}, demonstrating that perspective-trained metric depth models perform poorly on fisheye data. While undistorted camera representations sacrifice significant FoV or severs interpolating artifacts, applying a virtual focal length \( \frac{1}{f_{\text{virtual}}} = \tan\left(\frac{\pi}{H_{\text{erp}}}\right) \) to raw fisheye images or their ERP conversions results in even greater performance degradation. To ensure a fair comparison between DAC and pre-trained perspective models, we apply ERP conversion during fisheye testing for the perspective models as well, given that neither representation—raw fisheye nor ERP—falls within their original camera domain.

\begin{table*}[t!]
\caption{\textbf{Pretrained model performance on various representations of KITTI 360 dataset~\cite{journals/pami/LiaoXG23/kitti360}} 
}
\begin{center}
\begin{footnotesize}
\setlength\tabcolsep{0.1cm}
\begin{tabular}{l | l | l | l l l l  l  l }
\hline 
\textbf{Representation} & \textbf{Methods} & \textbf{Train Dataset} & $\pmb{\delta_1}$  $\uparrow$ & $\pmb{\delta_2}$  $\uparrow$ & $\pmb{\delta_3}$  $\uparrow$  & \textbf{Abs Rel}$\downarrow$ & \textbf{RMSE}$\downarrow$ & \textbf{log10}$\downarrow$  \\
\hline
\multirow{2}{*}{KITTI 360 Raw (FOV 180)} & Metric3Dv2~\cite{journals/corr/abs-2404-15506/metric3dv2} & Mix 16M & 0.7421 & 0.9498 & 0.9829 & 0.1679 & 3.0873 & 0.0739\\

& Metric3Dv2~\cite{journals/corr/abs-2404-15506/metric3dv2} & Outdoor 130K & 0.6400 & 0.9077 & 0.9763 & 0.1884 & 3.5698 & 0.0902\\


\hline

\multirow{2}{*}{KITTI 360 ERP (FOV 180)} &
Metric3Dv2~\cite{journals/corr/abs-2404-15506/metric3dv2} & Mix 16M & 0.7159 & 0.9323 & 0.9770 & 0.1997 & 4.5769 & 0.0811\\

& Metric3Dv2~\cite{journals/corr/abs-2404-15506/metric3dv2} & Outdoor 130K & 0.7675 & 0.9370 & 0.9756 & 0.1521 & 4.6610 & 0.0723\\


\hline

\multirow{2}{*}{KITTI 360 UD FoV 90} &
Metric3Dv2~\cite{journals/corr/abs-2404-15506/metric3dv2} & Mix 16M & 0.7581 & 0.9533 & 0.9738 & 0.1652 & 2.1454 & 0.0799\\

& Metric3Dv2~\cite{journals/corr/abs-2404-15506/metric3dv2} & Outdoor 130K & 0.8099 & 0.9582 & 0.9807 & 0.1469 & 2.1203 & 0.0650\\


\hline

\multirow{2}{*}{KITTI 360 UD FoV 120} &
Metric3Dv2~\cite{journals/corr/abs-2404-15506/metric3dv2} & Mix 16M & 0.6398 & 0.9285 & 0.9717 & 0.1929 & 2.3375 & 0.0968\\

& Metric3Dv2~\cite{journals/corr/abs-2404-15506/metric3dv2} & Outdoor 130K & 0.6635 & 0.9019 & 0.9685 & 0.1865 & 2.5982 & 0.0929\\


\hline

\multirow{2}{*}{KITTI 360 UD FoV 150} &
Metric3Dv2~\cite{journals/corr/abs-2404-15506/metric3dv2} & Mix 16M & 0.4840 & 0.8533 & 0.9551 & 0.2311 & 2.8692 & 0.1210\\

& Metric3Dv2~\cite{journals/corr/abs-2404-15506/metric3dv2} & Outdoor 130K & 0.4565 & 0.7788 & 0.9041 & 0.2498 & 3.2509 & 0.1355\\


\hline

\end{tabular}
\end{footnotesize}
\end{center}
\label{tab:metric:depth:zero-shot:distortion}
\end{table*}








\section{On Applying Camera Distortion Models}
\label{sec:distortion:model}

As described in Sec. \ref{sec:im2erp}, the conversion between actual image and the ERP can seamlessly handle different distortion models. In this section, we illustrate how we apply to two typical fisheye models: KB (OpenCV Fisheye) model~\cite{kannala2006generic} and MEI model~\cite{mei2007single}.

\subsection{KB Model}

KB model typically includes distortion parameters $k_1, k_2, k_3, k_4$.
Applying KB model to our Eq.~\ref{eq:distortion} can start from mapping our definition in Eq.~\ref{eq:gp:x} and Eq.~\ref{eq:gp:y} to the original KB model notations to get:
\begin{align}
    a & =x_t, \quad b = y_t\\
    r &= \sqrt{x_t^2 + y_t^2} \\
    \theta &= \arctan(r) = c
\end{align}
However, the direct use of $(x_t, y_t)$ can face numerical issue when the FOV is near $180^\circ$, when the dividing of $\cos c$ approaches 0 in computing them. A more numerical stable version supporting KB at $180^\circ$ is to use the numerators in Eq.~\ref{eq:gp:x} and Eq.~\ref{eq:gp:y}, denoted as $(\bar{x}, \bar{y})$. Then we can rewrite:
\begin{align}
    a & =\bar{x}, \quad b = \bar{y}\\
    r &= \sqrt{\bar{x}^2 + \bar{y}^2} \\
    \theta & = c
\end{align}
where we can keep the ratios $\frac{a}{r}, \frac{a}{r}$ consistent between two approaches, while avoiding numeric issues caused by dividing $\cos 0$.

The remaining process is exactly the same as the original KB model. \textbf{Fisheye distortion} is applied as: 
\begin{equation}
    \theta_d = \theta (1 + k_1 \theta^2 + k_2 \theta^4 + k_3 \theta^6 + k_4 \theta^8)
\end{equation}
The distorted point coordinates are \([x', y']\) where
\begin{align}
    x_d &= \left( \frac{\theta_d}{r} \right) a \\
    y_d &= \left( \frac{\theta_d}{r} \right) b
\end{align}
Finally, given a intrinsic model including $f_x, f_y, c_x, c_y, \alpha$ as parameters, the conversion into pixel coordinates \([u, v]\) can be written as:
\begin{align}
    u &= f_x (x_d + \alpha y_d) + c_x \\
    v &= f_y y_d + c_y
\end{align}

\subsection{MEI Model}

MEI model is general more complex by including parameters $\xi, k_1, k_2, p_1, p_2$, where an additional shift parameter $\xi$ is applied so that the model handle even larger FOV camera, and $p_1, p_2$ are including tangential distortion. 

Mapping our definitions to MEI model is even simpler. Note that $(\bar{x}, \bar{y}, \cos c)$ actually describe a point lying on the unit sphere, equalizing the Cartesian coordinates converted from the spherical coordinates. The projection coordinates \((p_u, p_v)\) are computed as:
\begin{align}
    p_u &= \frac{\bar{x}}{\cos c + \xi} \\
    p_v &= \frac{\bar{y}}{\cos c + \xi}
\end{align}
The distortion is then applied as:
\begin{equation}
    \rho^2 = p_u^2 + p_v^2
\end{equation}
\begin{align}
    p_u &\gets p_u \cdot (1 + k_1 \rho^2 + k_2 \rho^4) \\
    p_v &\gets p_v \cdot (1 + k_1 \rho^2 + k_2 \rho^4)
\end{align}
Tangential distortion is further applied as:
\begin{align}
    x_d &\gets p_u + 2 p_1 p_u p_v + p_2 (\rho^2 + 2 p_u^2) \\
    y_d &\gets p_v + p_1 (\rho^2 + 2 p_v^2) + 2 p_2 p_u p_v
\end{align}
The later projection is applied the same way as KB model.


\section{Efficient Up-Projection from Distorted Cameras via Lookup Table Approximation}

Up-projection is a crucial step to convert predicted depth maps into 3D point clouds. For perspective or ERP images, this process is straightforward, as the 3D ray associated with each pixel can be computed in closed form. However, up-projection from fisheye depth maps poses challenges due to the need to invert the distortion model, often requiring the solution of a high-order polynomial equation for each pixel based on the distortion parameters. This process is computationally expensive and impractical for real-time applications. 

Fortunately, pre-computed lookup tables can address this issue efficiently. These tables store a mapping from 2D image coordinates to 3D ray directions, allowing for real-time up-projection, which can be written as:
\begin{equation}
    \mathbf{L} : \mathbb{R}^2 \to \mathbb{R}^3, \quad \mathbf{L}(\mathbf{u}) = \mathbf{r}, \label{eq:lookup}
\end{equation}
where \(\mathbf{L}\) represents the lookup table, \(\mathbf{u} = (u, v) \in \mathbb{R}^2\) denotes the 2D image coordinates, and \(\mathbf{r} = (x, y, z) \in \mathbb{R}^3\) represents the corresponding 3D ray direction. The lookup tables can be generated using tools like OpenCV with gradient-based numerical methods or through simpler grid search approaches when tangential distortion parameters are negligible~\cite{journals/pami/LiaoXG23/kitti360}. in this work, we use similar grid search approach to computed lookup tables for Scannet++~\cite{conf/iccv/YeshwanthLND23/scannet++} based on their provided distortion and intrinsic parameters.

Notably, our DAC framework does not require approximated solutions for up-projection. In DAC, fisheye images are converted into ERP patches, which rely only on the forward distortion model. The resulting ERP depth maps can then be up-projected into 3D point clouds using each ERP coordinate's ray direction in a unit sphere, eliminating efficiency concerns. This represents a minor but valuable benefit of our approach.

Nevertheless, we identify two practical use cases for lookup tables in other contexts:
\begin{itemize}
    \item Visualization Purposes: Lookup tables efficiently map ERP patches and predicted ERP depth maps back to the original fisheye space for visualization, as illustrated in Fig.~\ref{fig:vis:main}. Specifically, ERP-to-image conversion for a fisheye image can also be performed efficiently using grid sampling, where each fisheye image coordinate is mapped to its floating-point location in the ERP space. The output of Eq.~\ref{eq:lookup} already provides tangent plane normalized coordinates, \(x_t = \frac{x}{z}\) and \(y_t = \frac{y}{z}\). Using the inverse of Gnomonic Geometry~\cite{weisstein:gnomonic}, the mapping to spherical coordinates \((\lambda, \phi)\) is derived as follows:
    \begin{align}
        \phi &= \sin^{-1} \left( \cos c \sin \phi_c + \frac{y_t \sin c \cos \phi_c}{\rho} \right) \\
        \lambda &= \lambda_c + \tan^{-1} \left( \frac{x_t \sin c}{\rho \cos \phi_c \cos c - y_t \sin \phi_c \sin c} \right)
    \end{align}
    where
    \begin{align}
    \rho &= \sqrt{x_t^2 + y_t^2} \nonumber\\ 
    c &= \tan^{-1} \rho \nonumber
    \end{align}
    However, this step is only needed for visualization purpose, not required for downstream tasks where up-projected 3D points are the most demanding. 
    \item Converting Z-Values to Euclidean Distances: For datasets like ScanNet++~\cite{conf/iccv/YeshwanthLND23/scannet++}, ground-truth depth maps recorded in Z-values must be converted to Euclidean distances for evaluation or inclusion in DAC training. This can be achieved efficiently using pre-computed ray directions from the fisheye's original incoming rays (not distorted by intrinsic parameters). The Euclidean distance for each pixel is calculated as: $D_{\text{Euclid}} = \frac{Z}{z}$, where \(Z\) represents the ground-truth Z-value, and \(z\) is the \(z\)-component of the ray direction \(\mathbf{r}\).
\end{itemize}

\section{Additional Visual Results}

\begin{figure*}[!tb]
    \centering
    \includegraphics[width=0.99\textwidth]{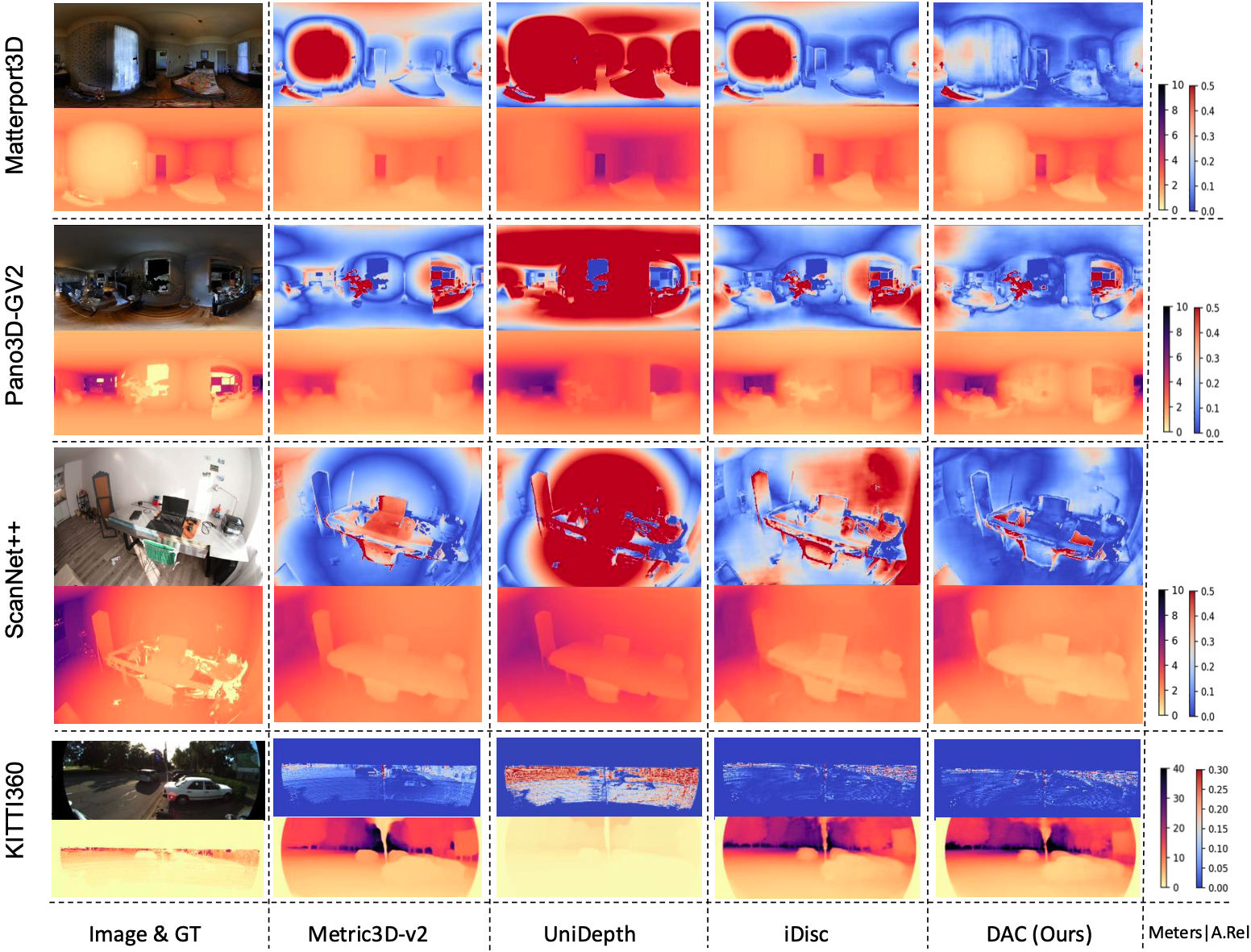}
    \caption{\textbf{Zero-Shot Qualitative Results.} For each dataset, an example is presented in two consecutive rows. The left column shows the original image and Ground-Truth depth map, followed by results from various methods. For each method, the top row displays the A.Rel map \( \downarrow \) and the bottom row shows the predicted depth map. The color range for depth and A.Rel maps is indicated in the last column.
    }
   \label{fig:vis:suplementary_qualitative_1}
\end{figure*}

\begin{figure*}[!tb]
    \centering
    \includegraphics[width=0.99\textwidth]{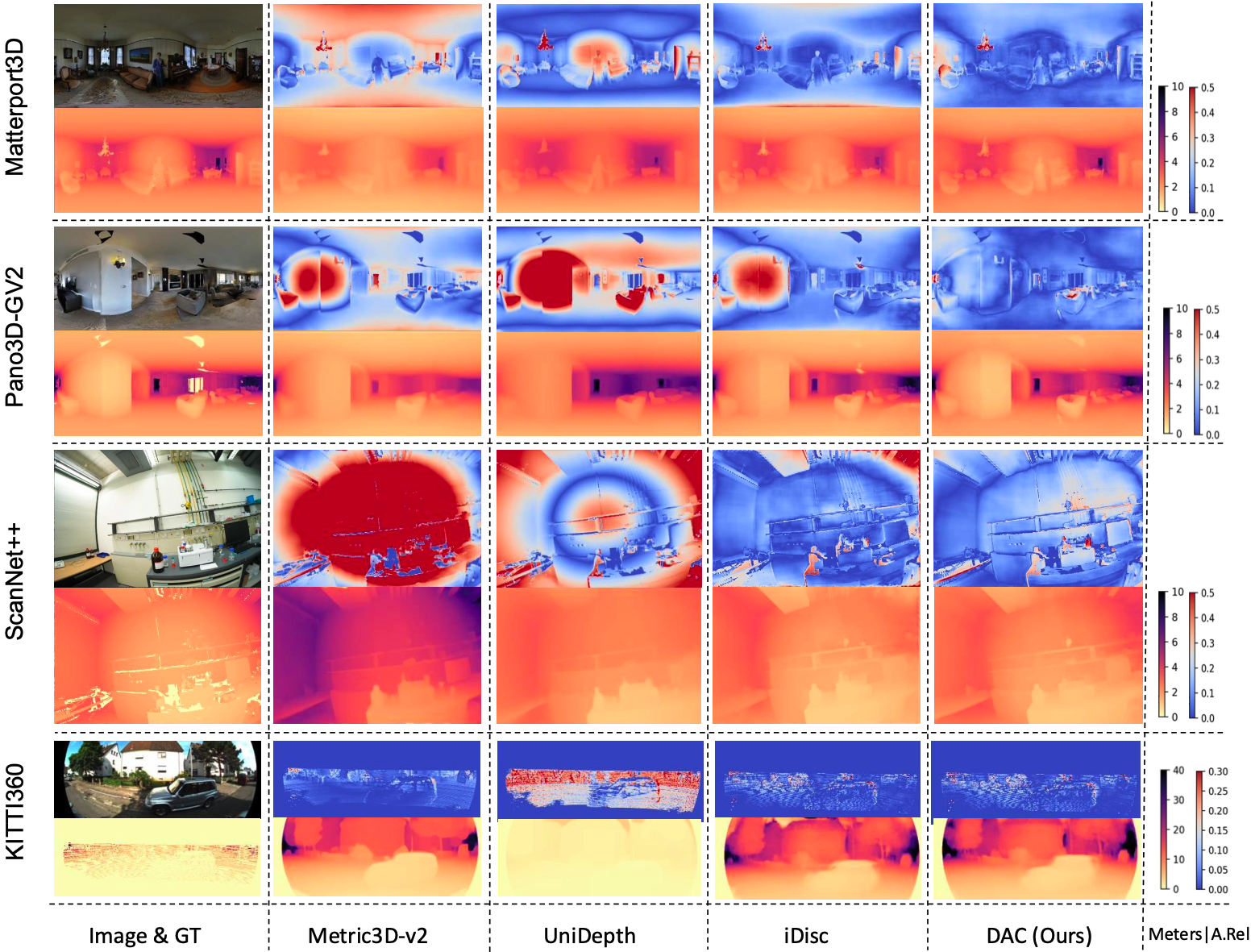}
    \caption{\textbf{Zero-Shot Qualitative Results.} For each dataset, an example is presented in two consecutive rows. The left column shows the original image and Ground-Truth depth map, followed by results from various methods. For each method, the top row displays the A.Rel map \( \downarrow \) and the bottom row shows the predicted depth map. The color range for depth and A.Rel maps is indicated in the last column.
    }
   \label{fig:vis:suplementary_qualitative_2}
\end{figure*}

\begin{figure*}[!tb]
    \centering
    \includegraphics[width=0.99\textwidth]{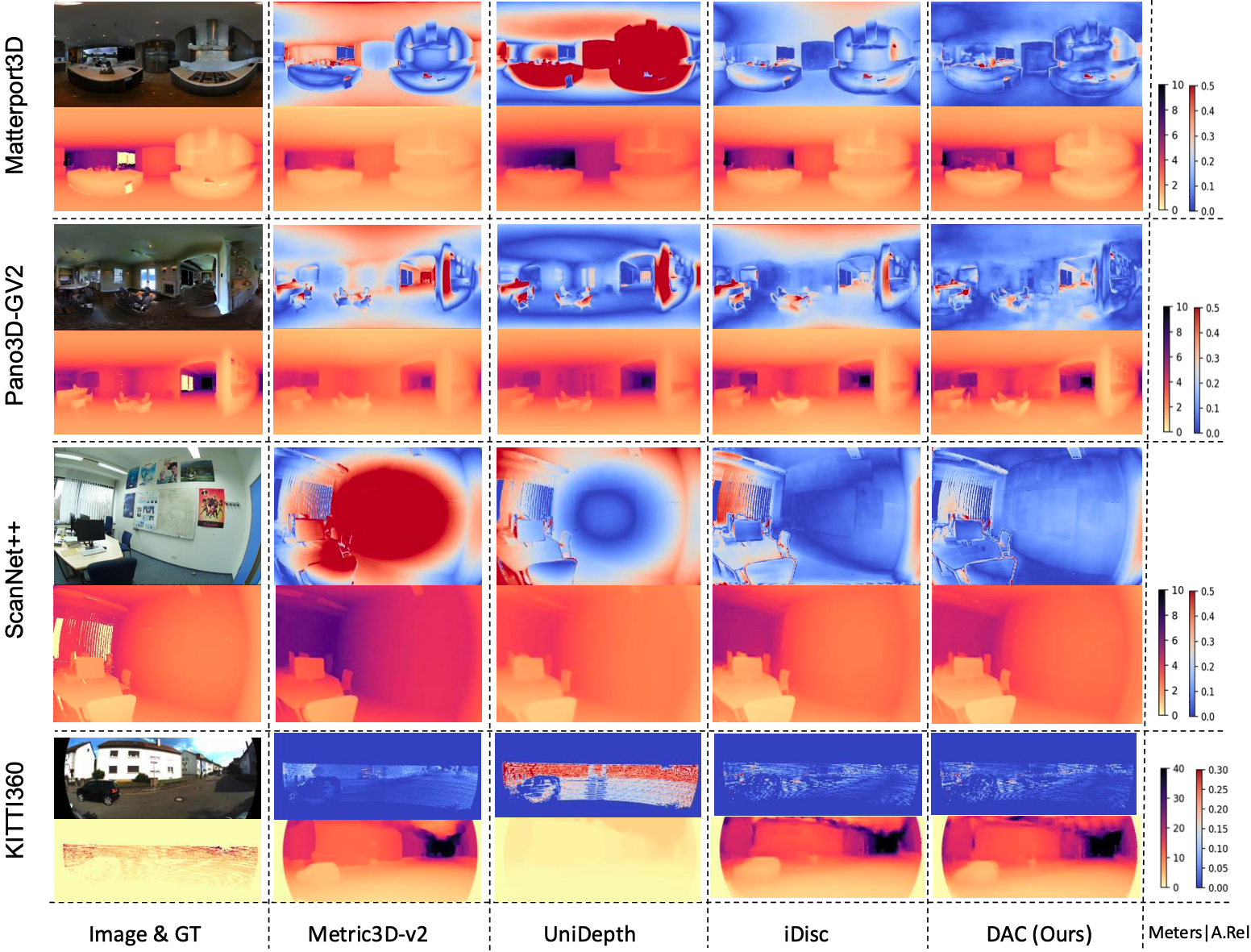}
    \caption{\textbf{Zero-Shot Qualitative Results.} For each dataset, an example is presented in two consecutive rows. The left column shows the original image and Ground-Truth depth map, followed by results from various methods. For each method, the top row displays the A.Rel map \( \downarrow \) and the bottom row shows the predicted depth map. The color range for depth and A.Rel maps is indicated in the last column.
    }
   \label{fig:vis:suplementary_qualitative_3}
\end{figure*}

In this section, we provide three additional set of visual comparisons of the competing methods on each large-FoV test set, namely: Matterport3D \cite{Matterport3D}, Pano3D-GV2 \cite{conf/cvpr/AlbanisZDGSAZD2/pano3d}, Scannet++ \cite{conf/iccv/YeshwanthLND23/scannet++}, and KITTI360 \cite{journals/pami/LiaoXG23/kitti360}, as shown in Fig. \ref{fig:vis:suplementary_qualitative_1}, \ref{fig:vis:suplementary_qualitative_2}, \ref{fig:vis:suplementary_qualitative_3}. Compared to Fig.~\ref{fig:vis:main}, visual results of Unidepth \cite{conf/cvpr/PiccinelliYSSLG24/unidepth} are also included for comparison.

Through visual comparisons, our DAC framework demonstrates sharper boundaries in the depth maps and more visually consistent scale in the depth visualization results. As seen in the A.Rel maps wrt. the ground-truth depth, our framework exhibits a significant advantage over each previous state-of-the-art method.

\end{document}